\documentclass{article} % For LaTeX2e
\usepackage{iclr2021_conference,times}

\usepackage{graphicx,subfigure}
\usepackage{comment}
\usepackage{amsmath,amssymb} % define this before the line numbering.
\usepackage{color}
\usepackage{bbm}
\usepackage{listings}
\usepackage{multirow}
\usepackage{xcolor}
\usepackage{float}
\usepackage{caption}

\newcommand{\ud}{\,\mathrm{d}}

\newcommand{\ip}[3]{\left< {#1}, {#2} \right>_{#3}}
\newcommand{\inner}[3]{\left\langle #2, #3\right\rangle_{_{\!\!H^#1}}}

\newcommand{\avg}[1]{\overline{#1}}

\newcommand{\cut}[1]{}
\newcommand{\der}[2]{\frac{\ud {#1}}{\ud {#2}}}
\newcommand{\pder}[2]{\frac{\partial {#1}}{\partial {#2}}}

\newcommand{\R}{\mathbb{R}}
\DeclareMathOperator*{\argmax}{arg\,max}

\newtheorem{defn}{Definition}

\usepackage[ruled]{algorithm}
\usepackage{algorithmicx}
\usepackage{algpseudocode}
\usepackage{multirow}

\usepackage[utf8]{inputenc} % allow utf-8 input
\usepackage[T1]{fontenc}    % use 8-bit T1 fonts
\usepackage{hyperref}       % hyperlinks
\usepackage{url}            % simple URL typesetting
\usepackage{booktabs}       % professional-quality tables
\usepackage{amsfonts}       % blackboard math symbols
\usepackage{nicefrac}       % compact symbols for 1/2, etc.
\usepackage{microtype}      % microtypography

\usepackage{hyperref}
\usepackage{url}

\title{Channel-Directed Gradients for Optimization of Convolutional Neural Networks}

\iclrfinalcopy

\author{Dong Lao, Peihao Zhu, Peter Wonka, Ganesh Sundaramoorthi\\
King Abdullah University of Science and Technology\\
Thuwal, Saudi Arabia\\
\texttt{\{dong.lao,peihao.zhu,peter.wonka,ganesh.sundaramoorthi\}@kaust.edu.sa} \\
}

\begin{document}

\maketitle

\begin{abstract}
We introduce optimization methods for convolutional neural networks that can be used to improve existing gradient-based optimization in terms of generalization error. The method requires only simple processing of existing stochastic gradients, can be used in conjunction with any optimizer, and has only a linear overhead (in the number of parameters) compared to computation of the stochastic gradient. The method works by computing the gradient of the loss function with respect to output-channel directed re-weighted $\mathbb{L}^2$ or Sobolev metrics, which has the effect of smoothing components of the gradient across a certain direction of the parameter tensor. We show that defining the gradients along the output channel direction leads to a performance boost, while other directions can be detrimental. We present the continuum theory of such gradients, its discretization, and application to deep networks. Experiments on benchmark datasets, several networks and baseline optimizers show that optimizers can be improved in generalization error by simply computing the stochastic gradient with respect to output-channel directed metrics.
\end{abstract}

\section{Introduction}

Stochastic gradient descent (SGD) is currently the dominant algorithm for optimizing large-scale convolutional neural networks (CNNs) \cite{lecun1998gradient,simonyan2014very,he2016deep}. Although there has been large activity in optimization methods seeking to improve performance, SGD still dominates in large-scale CNN optimization in terms of its generalization ability. Despite SGD's dominance, there is still often a gap between training and real-world test accuracy performance in applications, which necessitates research in optimization methods to increase generalization accuracy.

\begin{figure}[t]
\centering
\hspace*{-0.2in}
\includegraphics[width=1.08\textwidth]{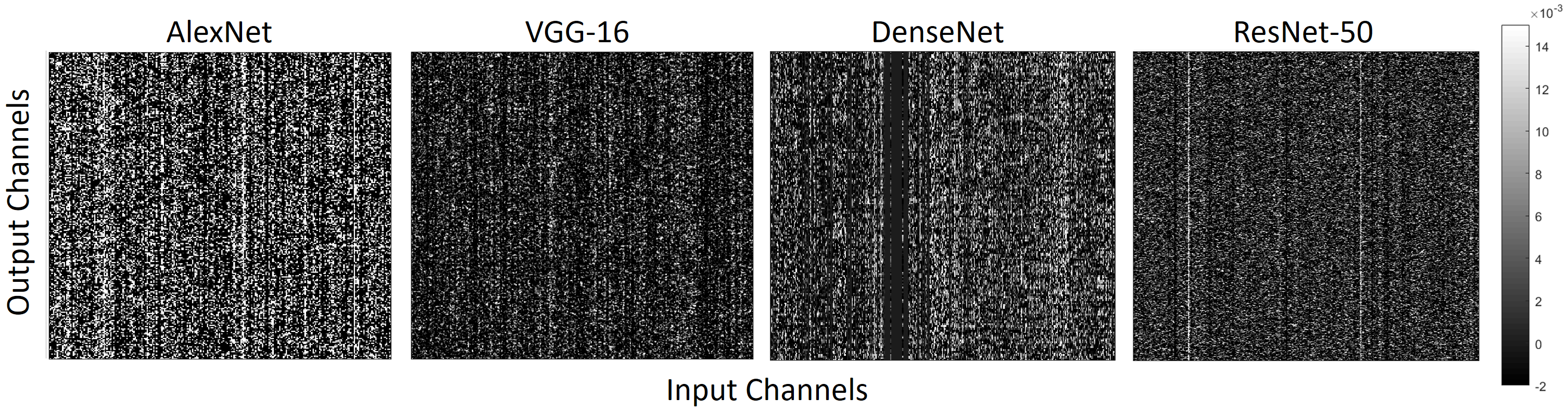}
\caption{Visualization of weights (parameter tensor) of convolutional layers trained on ImageNet using SGD. The vertical structures that indicate regularity (correlation) of the weights along the output channel direction. This pattern is frequently observed in layers. Our method is motivated by this empirical observation, and favors parameter correlations in this direction in optimization.}
  \label{fig:imagenet_tensor}
\vspace{-0.1in}
\end{figure}

In this paper, we observe that there is regularity in parameter tensors of learned CNN models, and we thus exploit regularity implicitly in optimization to derive new optimization methods that are simple modifications of traditional SGD to improve generalization. In particular, we empirically observe that parameter tensors in trained networks typically exhibit correlation over output channel dimension (see Figure~\ref{fig:imagenet_tensor}). We thus explore encoding correlation through smoothness in  optimization, which we show improves generalization accuracy as learning without imposing regularity may not fully learn it. We encode smoothness implicitly in stochastic gradient descent by considering new metrics on the parameter space of the network, and reformulating the notion of gradient. To do this, we treat the space of parameter tensors as a Riemannian manifold to derive gradients of the loss with respect to new metrics that promote regularity in the output channel dimension of the tensors by changing the geometry of the underlying space of tensors.

Our contributions are as follows. First, we formulate \emph{output channel-directed} Riemannian metrics (a re-weighted version of the standard $\mathbb{L}^2$ metric and another that is a Sobolev metric) over the space of parameter tensors. This encodes channel-directed regularity inherently in the gradient optimization without changing the loss. Second, we compute Riemannian gradients with respect to the metrics showing linear complexity (in the number of parameters) over standard gradient computation, and thus derive new optimization methods for CNN training. Finally, we apply the methodology to training CNNs and show the empirical advantage in generalization accuracy, especially with small batch sizes, over standard optimizers (SGD, Adam) on numerous applications (image classification, semantic segmentation, generative adversarial networks) with simple modification of existing optimizers.

\subsection{Related Work}

We briefly survey related work in deep network optimization; a detailed survey is \cite{bottou2018optimization}. Stochastic gradient descent (SGD), e.g., \cite{bottou2012stochastic}, samples a batch of data to tractably estimate the gradient of the loss function. As the stochastic gradient is a noisy version of the gradient, learning rates must follow a decay schedule in order to converge. Many methods have been formulated to choose learning rate over epochs and components of the gradient, including recent work on adaptive learning rates (e.g., \cite{duchi2011adaptive,zeiler2012adadelta,kingma2014adam,bengio2015rmsprop,loshchilov2017decoupled,luo2018adaptive}). For instance, Adam \cite{kingma2014adam} adaptively adjusts the learning rate so that parameters that have changed infrequently based on historical gradients are updated more quickly than parameters that have changed frequently. Another way to interpret such methods is that they change the underlying metric on the space on which the loss function is defined to an iso-tropically scaled version of the $\mathbb{L}^2$ metric given by a simple diagonal matrix; we change the metrics an-isotropically. We show that our method can be used in conjunction with such methods by simply using the stochastic gradient computed with our metrics to boost performance.

As the stochastic gradient is computed based on sampling, different runs of the algorithm can result in different local optima. To reduce the variance, several methods have been been formulated, e.g.,  \cite{defazio2014saga,johnson2013accelerating}. We are not motivated by variance reduction, rather we are motivated by inducing regularity in optimization to improve generalization. However, as our method effectively smooths the gradient, our empirical experiments do indicate reduced variance with our metrics compared to SGD.

Another method motivated by variance reduction is the recent work of \cite{osher2018laplacian}, where the stochastic gradient is pre-multiplied with an inverse Laplacian smoothing matrix. For CNNs, the gradient with respect to parameters is rasterized in row or column order of network filters before smoothing, which still lowers variance. Our work is inspired by \cite{osher2018laplacian}, though we are motivated by incorporating structured regularity of the parameter tensor inherently in the optimization. \cite{osher2018laplacian} can be interpreted as using the gradient of the loss with respect to a Sobolev metric. Our major insight over \cite{osher2018laplacian} is that keeping the multi-dimensional structure of the parameter tensor (rather than rasterizing) and preferentially defining the Sobolev metric with respect to the output-channel direction boosts generalization accuracy, while other directions appear to have no boost. Secondly, we introduce a re-weighted $\mathbb{L}^2$ metric that preferentially treats the output-channel direction, and can be implemented with one line of Pytorch code, is linear (in parameter size) complexity, and achieves similar results (in many cases) to our channel-directed Sobolev metric, boosting generalization of SGD and Adam. Third, our channel-directed Sobolev gradient can be implemented in linear cost rather than quasi-linear (not requiring FFT to compute). Sobolev gradients have been used in computer vision \cite{sundaramoorthi2007sobolev,charpiat2007generalized} for their coarse-to-fine evolution properties \cite{sundaramoorthi2008coarse} and we adapt that formulation to channel-directed metrics for CNNs.

We formulate Sobolev gradients by considering the space of parameter tensors as a Riemannian manifold, and choosing the metric (i.e., inner product) on the tangent space to be a Sobolev metric. By choosing a metric, gradients intrinsic to the manifold can be computed and gradient flows decrease loss. Other Riemannian metrics have been used for optimization in machine learning, e.g., \cite{amari1998natural,marceau2016practical,hoffman2013stochastic,gunasekar2018implicit,gunasekar2020mirrorless} and tangentially relate to our work. These works are based on Amari's \cite{amari1998natural} information geometry on probability measures, and the metric considered is the Fisher information metric. The motivation for these methods is re-parametrization invariance of optimization, whereas our motivation is imposing regularity directly in the parameter space. Most of these methods relate to density estimation as the metric is on probability measures. \cite{gunasekar2018implicit} notes that even vanilla gradient descent has certain implicit bias relating to the underlying metric on the space. In \cite{gunasekar2020mirrorless} the Hessian metric (in the convex case) is analyzed and relates to mirror descent. These metrics are data-dependent and the gradient is challenging to compute, requiring (a large) inverse matrix computation. Moreover, they do not exploit the channel dimension regularity, the main purpose of our work.

\section{Channel-Directed Gradients}

We now present the theory to define channel-directed gradients. To do this, we formulate new metrics on the space of tensors, and then derive analytic formulas for channel-directed gradients in terms of the standard $\mathbb{L}^2$ gradient. As we show, our channel-directed gradients effectively smooth the components of the $\mathbb{L}^2$ gradient across a certain direction of the parameter tensors of the CNN. Another interpretation is we are changing the geometric structure of the loss landscape (without changing the loss) to a more smooth one by changing the underlying metric of the space on which the loss is defined.

Our metrics are motivated by the empirical observation that a certain dimension of parameter tensors in trained deep networks exhibits regularity (see Figure~\ref{fig:imagenet_tensor}), and thus our method exploits this regularity implicitly in optimization. If we visualize the parameter tensor along the input and output dimensions of the parameter tensor, we see mostly what appears to be random noise, however, there are in addition, regular (correlated) patterns along the output channel direction, implying that each output channel of a layer of a network uses similar (regularly-varying) weightings of input channels. Our metrics thus favor that gradient updates during optimization exhibit correlation, which we show in experiments leads to optimization that generalizes better.

\subsection{Background on  Riemannian Gradients}

We first briefly present the definition of gradient on a Riemannian manifold, and show the explicit dependence of the gradient on the chosen metric on the manifold. More detailed theory can be found in \cite{carmo1992riemannian,abraham2012manifolds}. We note that a manifold $\mathcal X$ is a space that is locally linear around each point $X\in \mathcal X$ in the space, and the linear space at each point is called the \emph{tangent space}, denoted $T_X \mathcal X$. A \emph{Riemannian manifold}, in addition, has a smoothly varying positive definite bilinear form $\left< \cdot,\cdot \right>$ (called the \emph{metric}) on the tangent space. This metric allows one to define the notion of lengths of curves on the space, in addition to many other operations, including gradients of functions defined on the space.
\begin{defn}[Gradient of a Function]
    Let $\mathcal X$ be a Riemannian manifold, and $f : \mathcal X \to \R$ be a function. The directional derivative of $f$ at $X\in \mathcal X$ along a direction $k\in T_X \mathcal X$ is defined as $\ud f(X)\cdot k = \der{}{\varepsilon} \left. f(X+ \varepsilon k) \right|_{\varepsilon = 0}$. The {\bf gradient} of $f$ at $X\in \mathcal X$ is the vector, $\nabla f(X) \in T_{X}\mathcal X$, that satisfies the relation 
    \begin{equation}
        \ud f (X) \cdot k = \ip{\nabla f(X)}{ k }{},\,\, \mbox{for all } k\in T_X \mathcal X.
    \end{equation}
\end{defn}
From the definition, we note that ``the'' gradient will depend on the choice of the metric on the manifold. We note that any such gradient will decrease the the function $f$ by moving infinitesimally in the tangent space in the direction of negative the gradient as $\ud f(X) \cdot k = -\| \nabla f(X) \|^2 < 0$ when $k=-\nabla f(X)$, where $\|\cdot\|$ is the norm induced by $\left< \cdot, \cdot  \right>$. The gradient flow, defined by the differential equation $\dot{X}_t = -\nabla f(X_t)$, will converge to a local minimum. In our application of this theory to CNN optimization, $f$ will be the loss function, and $\mathcal X$ will be the space of parameter tensors. In this case, as the tensor is multi-dimensional, the gradient flow will be a partial differential equation.

A consequence of this definition is that the gradient is the direction (up to a scale factor) in the tangent space that optimizes the following problem:
\begin{equation}
    \argmax_{k\in T_X \mathcal X \backslash \{0\} } \frac{|\ud f(X)\cdot k|}{\|k\|}.
\end{equation}
Thus, the gradient can be regarded as the most efficient direction as it maximizes the ratio of the change in energy by perturbing in a direction $k$ over the cost (defined by the metric) of $k$. Thus, by constructing the metric to have small costs for perturbations (directions) that we prefer for gradients, the gradient flow will move in these preferential directions while minimizng the fucntion, and thus land in more favorable local minima. In particular, we construct metrics that favor gradients with output channel direction regularity and this does induce regularity in the final tensor.

\subsection{Channel-Directed Metrics}

As gradients of a function depend on the metric structure on the underlying space, we re-define the metric on the underlying space so that tensors that differ smoothly along the output channel direction have small distance. In existing deep network gradient-based optimization schemes, the underlying metric on the loss function is assumed to be the standard Euclidean $\mathbb{L}^2$ metric. We will consider a re-weighted version of the $\mathbb{L}^2$ metric and Sobolev metrics that favor regularity in the output channel direction of parameter tensors. To formulate the methodology, we start from a continuum formulation, where we treat weight tensors in the continuum, formulate the metrics in the continuum and then in the next sub-section derive the gradients with respect to these metrics. Finally, we discretize gradient flows in the implementation to derive iterative schemes.

Let $X : \mathcal{O} \times \mathcal{I} \times \mathcal{H} \times \mathcal{W} \to \R$ denote a parameter tensor of a deep network (from a layer of a convolutional network).  Here $\mathcal{O}=[0,O]$ denotes indices to the output channel dimension of the tensor, $\mathcal{I}=[0,I]$ denote the indices to the input channel, and $\mathcal{H}=[0,H], \mathcal{W}=[0,W]$ denote indices to the height and width dimension of the spatial filters of the tensor. The metric is defined on the tangent space to the space of such $X$. An element of the tangent space will have the same form of the tensor, i.e., $k : \mathcal{O} \times \mathcal{I} \times \mathcal{H} \times \mathcal{W} \to \R$. The $\mathbb{L}^2$ (called $H^0$ from now on) metric is defined as 
\begin{equation}
\inner{0}{k_1}{k_2} = \int_{\mathcal{O},\mathcal{I},\mathcal{H},\mathcal{W}} k_1(o,i,h,w) \cdot k_2(o,i,h,w)  \ud o \ud i \ud h \ud w,
\end{equation}
where $k_1, k_2$ are in the tangent space of tensors. We now define a re-weighted version of $H^0$ that favors tangent vectors that have global smoothness in the direction of the $\mathcal{O}$ dimension:
\begin{equation} \label{eq:reweightedH0_metric}
    \ip{k_1}{k_2}{H^0_{\lambda}} = \int_{\mathcal{I},\mathcal{H},\mathcal{W}} \bar{k}_1(i,h,w)\cdot\bar{k}_2(i,h,w) \ud i \ud h \ud w  + \frac{\lambda}{O} \inner{0}{k_1-\bar{k}_1}{k_2-\bar{k}_2},
\end{equation}
where $\lambda>0$ is a hyper-parameter, and $\bar{k}$ is the average value in the output channel direction, i.e.,

\begin{equation}
\bar{k}(i,h,w) = \frac{1}{O}\int_{\mathcal{O}} k(o,i,h,w) \ud o.
\end{equation}
The metric in \eqref{eq:reweightedH0_metric} splits the tangent vector into global translations in the output channel direction and its orthogonal complement, i.e., the deformation. The weight $\lambda$ is used to control the weighting between the translation and deformation components, i.e., larger values of $\lambda$ means that deformations more heavily influence the norm of the perturbation. As shown in the next sub-section that means gradients with respect to this metric have higher weighted channel-directed translations than deformations.

Next, we introduce channel-directed versions of the Sobolev metric, defined as follows:
\begin{align}
\inner{1}{k_1}{k_2} &= \frac{1}{O}\inner{0}{k_1}{k_2} + \lambda O \inner{0}{\pder{k_1}{o}}{\pder{k_2}{o}} \\
\ip{k_1}{k_2}{\tilde H^1} &= \int_{\mathcal{I},\mathcal{H},\mathcal{W}} \bar{k}_1(i,h,w)\cdot \bar{k}_2(i,h,w) \ud i \ud h \ud w + 
\lambda O \inner{0}{\pder{k_1}{o}}{\pder{k_2}{o}}, \label{eq:metric_H1}
\end{align}
where $\pder{}{o}$ indicates the partial derivative with respect to the the output channel direction. The partial derivative in the $o$-direction implies that tensor perturbations that are smooth along the $o$-direction are close with respect to these metrics, which will imply that the corresponding gradients will exhibit smoothness in this direction, i.e., convolution filters that are nearby in the output direction will exhibit correlation. The first metric is analogous to the usual Sobolev metric being a weighted combination of the $H^0$ metric and the $H^0$ metric of the derivative, except only considering the derivative with respect to one direction. The second metric is similar to the first except that we use the $H^0$ metric of the channel-directed average rather than that of the perturbation itself. As we will see, both have similar properties, but the latter is computationally less costly to compute. The scale factors of $O$ in the expressions above are so that the metric is scale invariant with respect to different sizes of output channels. The part of the metric with the partial derivative component implies that tensors that differ in the output channel direction by a non-smooth perturbation are far away in distance. In the latter metric, tensors that differ by just a channel-directed translation are close. Compared with the re-weighted $H^0$ metric \eqref{eq:reweightedH0_metric}, the latter Sobolev metric promotes smooth perturbations beyond global translations.

We have presented only channel-directed metrics that preferentially treat the output channel dimension of the tensor as our empirical experiments demonstrate that promoting regularity in other directions is detrimental to optimization performance.

\subsection{Computing Channel-Directed Gradients}

\begin{figure}[t]

\centering
  \begin{minipage}[t]{0.45\textwidth}
    \includegraphics[width=0.95\textwidth]{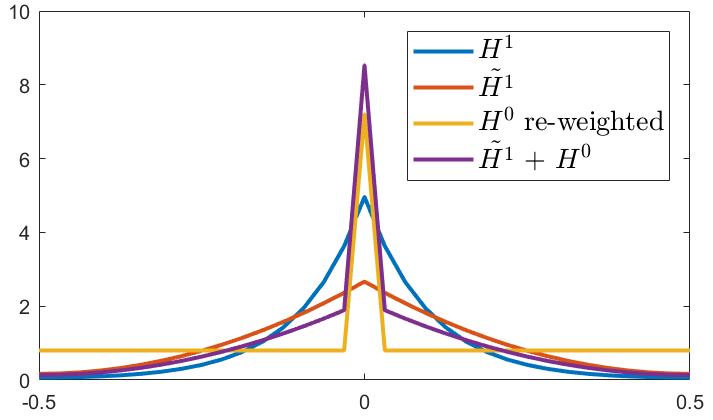}
  \end{minipage}\hfill
  \begin{minipage}[b]{0.55\textwidth}

\caption{{\bf Visualization of kernels applied to the $H^0$ gradient under different metrics for $\lambda=1$.} This illustrates the smoothing effect of the metrics. In computation, linear cost formulas are applied to compute the gradients not using the convolution interpretation.}\label{fig:kernels}
  \end{minipage}

\end{figure}

\begin{comment}
\begin{figure}[t]
\centering
\includegraphics[width=0.4\textwidth]{Figures/30.png}
             \vspace{-0.1in}
  \caption{{\bf Visualization of kernels applied to the $H^0$ gradient under different metrics for $\lambda=1$.} This illustrates the smoothing effect of the metrics. In computation, linear cost formulas are applied to compute the gradients not using the convolution interpretation.}
  \label{fig:kernels}
\end{figure}
\end{comment}

We now compute gradients with respect to the metrics defined in the previous sub-section in terms of the $H^0$ gradient so that simple processing of the existing gradient can be done with no other modification of existing optimization code. To compute the relation between the channel-directed gradients and the usual $H^0$ gradient, we note the relation between the directional derivative of a loss function, the gradient and the metric:
\begin{equation}\label{eq:gradient}
    \ud L (X)\cdot k = \inner{0}{\nabla_{H^0} L(X)}{k} = 
    \ip{\nabla_{H^0_{\lambda}} L(X)}{k}{H^0_{\lambda}} = 
    \inner{1}{\nabla_{H^1} L(X)}{k} =  \ip{\nabla_{H^1} L(X)}{k}{\tilde H^1},
\end{equation}
where $d L(X)\cdot k = \lim_{\varepsilon\to 0} \frac{ L(X+\varepsilon k) - L(X) }{ \varepsilon }$ is the directional derivative in the direction of the perturbation $k$. Note that the expression above indicates the directional derivative is equal to the inner product (metric) between the gradient with respect to the metric and the perturbation. This holds for any metric. With this relation, we may compute the channel-directed gradients in terms of the $H^0$ gradient. Derivations are in the Supplementary materials. Letting $f = \nabla_{H^0} L(X)$, we have

\begin{align}
    \label{eq:reweightedH0_grad}
    \nabla_{H^0_{\lambda}} L(X) &= \bar{f} + \frac{1}{\lambda} (f-\bar{f}) \\
    \label{eq:H1_grad}
    f = \nabla_{H^1} L(X) - \lambda O^2 \pder{^2}{o^2} \nabla_{H^1} L(X) &\quad \mbox{ and }\quad
    f = \avg{ \nabla_{\tilde H^1} L(X)} - \lambda O^2 \pder{^2}{o^2} \nabla_{\tilde H^1} L(X),
\end{align}
where the last two expressions are second order ordinary differential equations (ODE), whose solution we discuss next. Notice that the re-weighted $H^0$ gradient \eqref{eq:reweightedH0_grad} simply re-weights the channel-directed translation component and the deformation component of the $H^0$ gradient differently, i.e., as $\lambda$ gets larger, the channel-directed translation becomes more prominent.

In obtaining the ODE expressions for the Sobolev gradients above, we have assumed periodic boundary conditions in the $\mathcal{O}$ dimension\footnote{Since ordering of filters along the channel direction in a CNN has no particular significance, choosing periodic or non-periodic boundary conditions is arbitrary.  Periodic conditions induces smoothness between the starting and ending filters in the $o$-dimension. The periodic condition enables a simpler computational solution.}. In this case, the Sobolev gradients can be interpreted as the circular convolution of the $H^0$ gradient with convolution kernels given as 
\begin{equation}
K(o) = \frac{\cosh\left[ \lambda^{-1/2}(o-0.5) \right] }{ 2\sinh\left[\lambda^{-1/2} \right]  }, \quad 
\tilde K(o) = 1 + \frac{o^2 - o + 1/6}{ 2\lambda }, \,\, \mbox{ for }\, o\in [0,1],
\end{equation}
for each of the $H^1$ and $\tilde H^1$ metrics, respectively, where $o$ above is scaled by $O$ to be between 0 and 1, and the circular convolution is given by 
\begin{equation}
    \nabla_{\tilde H^1} L(X)(o,i,h,w) = \frac{1}{O} \int_{\mathcal{O}} \tilde K( (o - \tilde{o})/ O ) f(\tilde{o},i,h,w) \ud \tilde{o}.
\end{equation}
Note that the re-weighted $H^0$ solution also has an interpretation of convolution with respect to a smoothing kernel. Figure~\ref{fig:kernels} shows plots of various kernels for the parameter $\lambda$ chosen in experiments. For each $o$, the Sobolev or re-weighted $H^0$ is a local average whose weights die far away from $o$. Thus, we can see that the effect of the metrics is to induce smoothness of the gradient along the output channel direction.

The second version of the Sobolev gradient need not use the convolution formula for computation, as one can just integrate the ODE twice (after noting that the channel-directed averages for both the $H^0$ and Sobolev gradient are the same). This saves one from having to compute the convolution directly, and hence a reduction in computational cost from quadratic (or quasi-linear with an FFT) to linear in $O$ given the $H^0$ gradient. The second version of the Sobolev gradient can be computed as
\begin{align}
    \label{eq:tildeH1_linear_soln_1}
    g(o,i,h,w) &= g(0,i,h,w) + o \pder{g}{o}(0,i,h,w) - \frac{1}{\lambda} \int_0^{o} (o-\tilde{o})(f(oO,i,h,w)-\bar{f}(i,h,w)) \ud \tilde o \\
     \label{eq:tildeH1_linear_soln_2}
    \pder{g}{o}(0,i,h,w) &= -\frac{1}{\lambda} \int_0^1 o(f(oO,i,h,w) - \bar{f}(i,h,w) ) \ud o \\
     \label{eq:tildeH1_linear_soln_3}
    g(0,i,h,w) &= \int_0^1 \tilde K(o) f(oO, i, h, w ) \ud o, \quad o\in [0,1]
\end{align}
where $g = \nabla_{\tilde H^1} L(X)$ is the second version of the Sobolev gradient and $f=\nabla_{H^0} L(X)$, which are just three integrals that can be computed in linear complexity with respect to $O$.
The gradient flows under these metrics are given by 
\begin{equation} \label{eq:gradient_flow}
   \dot{X}_t = -\nabla L(X_t),
\end{equation}
where $t$ denotes the artificial time variable, $\dot{X}$ is the time derivative of the parameter tensor, and $\nabla$ denotes the gradient with respect to the desired metric. Under any metric, this reduces the loss.

\subsection{Properties of Channel Directed Gradient Flows}
We describe some properties of the resulting gradient flows according to the metrics defined in the previous sections.

{\bf Coarse-to-Fine Evolution and Removal of Some Local Minima}: In \cite{sundaramoorthi2008coarse}, it is shown that gradient flows with respect to Sobolev metrics evolve in a coarse-to-fine fashion, deforming according to coarse-scale perturbations before moving to finer scale perturbations. This can avoid being trapped in local minima due to fine-scale structures. It is also shown that when we change the metric on the space $\mathcal X$, the loss landscape changes and some local minima with respect to $H^0$ may change to other critical points with respect to Sobolev, and numerically some local minima may cease to exist. That is, local minima due to fine-scale structures can be removed by switching to a Sobolev metric. As local minima that are from wide, flat minima generalize well, the removal of local minima due to localized fine structures may encourage convergence at wider, flat minima and hence generalize better than ordinary SGD.

{\bf Regularity of the Weight Tensor}: By the convolution formulas above, we can see that the Sobolev gradients are a smoothing of the usual $H^0$ gradient. Noting that the gradient flow \eqref{eq:gradient_flow} integrates the (smooth) gradients over time, the final tensor will also be smooth in the output direction provided the initialization is smooth. In practice, in applications with deep networks, one typically initializes the weight tensor to be random noise, so the final tensor may exhibit some randomness, but the final tensor is sum of a smooth component and the randomness, and so it exhibits regularity, e.g., strong correlation across nearby output-channel components in the weight tensor, as we verify in experiments. Further, experiments (see Section~\ref{sec:experiments}) indicate that final results with respect to our metrics are less dependent on initialization than SGD, which suggests that the initial randomness may die out.

\section{Application to Stochastic Gradient Descent and Implementation}
To apply re-weighted $H^0$ and Sobolev channel-directed gradients to optimizing deep convolutional networks based on stochastic gradient descent or its variants, we discretize the gradient flow \eqref{eq:gradient_flow} according to the forward Euler method. We approximate the standard $H^0$ gradient of the loss, $\nabla_{H^0} L(X)$, using a mini-batch, as is standard in deep learning. We then use this approximation of the $H^0$ gradient to approximate the $\tilde H^1$ gradient, $\nabla_{\tilde H^1} L(X)$, by discretizing \eqref{eq:tildeH1_linear_soln_1}-\eqref{eq:tildeH1_linear_soln_3} using a standard Riemann sum. Note that \eqref{eq:tildeH1_linear_soln_1} can be computed for each $o$, the output channel index of the tensor, with the cumulative sum ({\tt\small CUMSUM}) operation, which is linear in cost, as are \eqref{eq:tildeH1_linear_soln_2} and \eqref{eq:tildeH1_linear_soln_3}. We compute the Sobolev gradient for each convolutional layer parameter tensor independent of others. We use $\lambda=1$ for $\tilde H^1$ gradient and add it to a scaled version (by a hyper parameter) of the $H^0$ gradient (as shown in Figure~\ref{fig:kernels}) to avoid over-smoothing. The re-weighted $H^0$ gradient is computed by using \eqref{eq:reweightedH0_grad} from the $H^0$ stochastic gradient approximation. Both gradients require few additional lines of code; the code for re-weighted $H^0$ is shown in Figure~\ref{fig:code}. Thus, the channel-directed gradients replace the usual one, and all other additions to standard SGD (e.g., momentum, Adam, etc) can be used as usual.

\cut{
Note that although all kernels are normalized in such a way that the integral of the kernel is equal to 1, in practice, this normalization can be absorbed into learning rate selection. Empirically, best result is obtained by keeping the deformation component with the same magnitude as original $H^0$, and using the same learning rate as SGD. \textcolor{red}{Not sure what this means}

\begin{center}
\includegraphics[width=\textwidth]{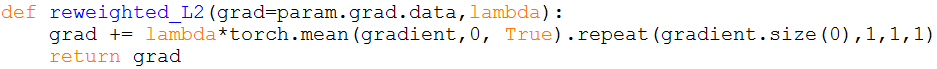}
\textcolor{red}{"grad" and "gradient" are same variable.  use also comments to specify what the input arguments are.  also you should use $H^0$ as same. can we use vertabtim so people can cut/paste code?}
\end{center}
}

\begin{figure}[t]
\vspace{-0.1in}
\begin{center}
\begin{lstlisting}[language=Python]
def reweighted_L2_grad(grad=param.grad.data,lambda):
    #grad: L2 gradient; lambda>0 weights translation of L2 grad
    grad += lambda*torch.mean(grad,0,True).repeat(grad.size(0),1,1,1)
    return grad
\end{lstlisting}
\end{center}
\vspace{-0.2in}
\caption{Pytorch code to compute the re-weighted $\mathbb{L}^2$ ($H^0_\lambda$) gradient from the $H^0$ gradient.} \label{fig:code}
\vspace{-0.1in}
\end{figure}

\section{Experiments} \label{sec:experiments}
We test our proposed channel-directed metrics with different baseline optimizers and tasks. Our intent is to show that any baseline method task can be improved just by switching the gradient with respect to channel directed metrics in the optimizer. We fix $\lambda = 1$ for channel-directed metrics unless specified otherwise. Table~\ref{Table:settings} shows the settings for each experiment. Experiments are run on a single NVIDIA Titan Xp GPU except for GANs, which are run on a Tesla v100 GPU due to memory requirements.

\begin{table}[H]
\fontsize{8}{10}\selectfont
  \vspace{-0.15in}
  \caption{{\bf Experimental settings.}}
  \label{Table:settings}
  \centering
  \begin{tabular}{lllllcc}
    \toprule
    Task&Dataset&Baseline&Network&Batch Size&Epochs&Initial LR\\
    \hline
    
    \multirow{5}{*}{Image Classification}&\multirow{4}{*}{Cifar-10}&\multirow{2}{*}{SGD}&ResNet-56&128,32,8&240&0.1\\
    &&&VGG-16&128,8,6&240&0.01\\
    &&ADAM&ResNet-56&128,32,8&200&1e-3\\
    &&LS&ResNet-56&128,32,8&240&0.1\\
    &MNIST&SGD&Two-layer Conv&100&100&0.01\\
    Semantic Segmentation&PascalVOC&SGD&ResNet50&2&70&7e-3\\
    Image Generation (GAN)&CityScapes&SGD&SPADE&2&100&1e-4,4e-4\\
    \bottomrule
  \end{tabular}
  \vspace{-0.1in}
\end{table}

{\bf Image Classification:}
We experiment on CIFAR-10~\cite{krizhevsky2009learning}. We test the combination of our channel-directed metrics with both SGD and ADAM on ResNet-56 \cite{He_2016} and VGG-16 \cite{simonyan2014very} following \cite{osher2018laplacian}'s settings. For SGD, we set the initial learning rate to be 0.1 and 0.01 on ResNet-56 and VGG-16 respectively with momentum 0.9 and weight decay 5e-4. For ADAM, we set the initial learning rate to 0.01. We decrease the learning rate by a factor of 10 every 40 epochs as \cite{osher2018laplacian}. Results presented in this section are the average of at least 10 independent trials.

\begin{table}[H]
\fontsize{8}{10}\selectfont
  \centering
\hspace*{-0.2in}
\caption{{\bf Test accuracy on CIFAR-10.} Channel-directed metrics improve $H^0$ in all cases. Best case, $>$10\% of error can be reduced by using our channel-directed metrics. Results average 10 trials.}
     \label{Table:cifar}
  \begin{tabular}{l|ccc|ccc|l|ccc}
    \toprule
    Architecture&\multicolumn{3}{c}{ResNet-56}&\multicolumn{3}{c|}{VGG-16}&Architecture&\multicolumn{3}{c}{ResNet-56} \\
    %\cmidrule(r){1-2}
    Batch size&128&32&8&128&8&6&Batch size&128&32&8\\
    \hline
    SGD&93.24&91.96&86.54&93.02&92.31&91.88&ADAM&91.20&91.04&89.53\\
    \hline
    $+\tilde{H^1}$&93.29&92.13&87.99&93.26&92.77&92.25&$+\tilde{H^1}$&91.42&91.13&90.02\\
    Error reduced\%&0.7\%&2.1\%&10.8\%&3.4\%&6.0\%&4.6\%&Error reduced\%&2.5\%&1.0\%&4.7\%\\
    \hline
    $+H^0_\lambda$&93.38&92.10&88.04&93.19&92.79&92.43&$+H^0_\lambda$&91.20&91.06&89.70\\
    Error reduced\%&2.1\%&1.7\%&11.1\%&2.4\%&6.2\%&6.8\%&Error reduced\%&0.0\%&0.2\%&1.6\%\\
    \bottomrule

  \end{tabular}
         \vspace{-0.1in}
\end{table}

Table~\ref{Table:cifar} shows the test accuracy under different settings. Both channel-directed metrics achieve improvement over $H^0$. A greater advantage over the baseline is achieved when the batch size is small as the stochastic gradient is noisy, and our method imposes regularity. In most cases, all channel-directed gradients perform similarly, but ${\tilde H^1}$ performs significantly better with ADAM.

\begin{figure}[t]
\centering
\minipage{0.45\textwidth}
\includegraphics[height=1.6in]{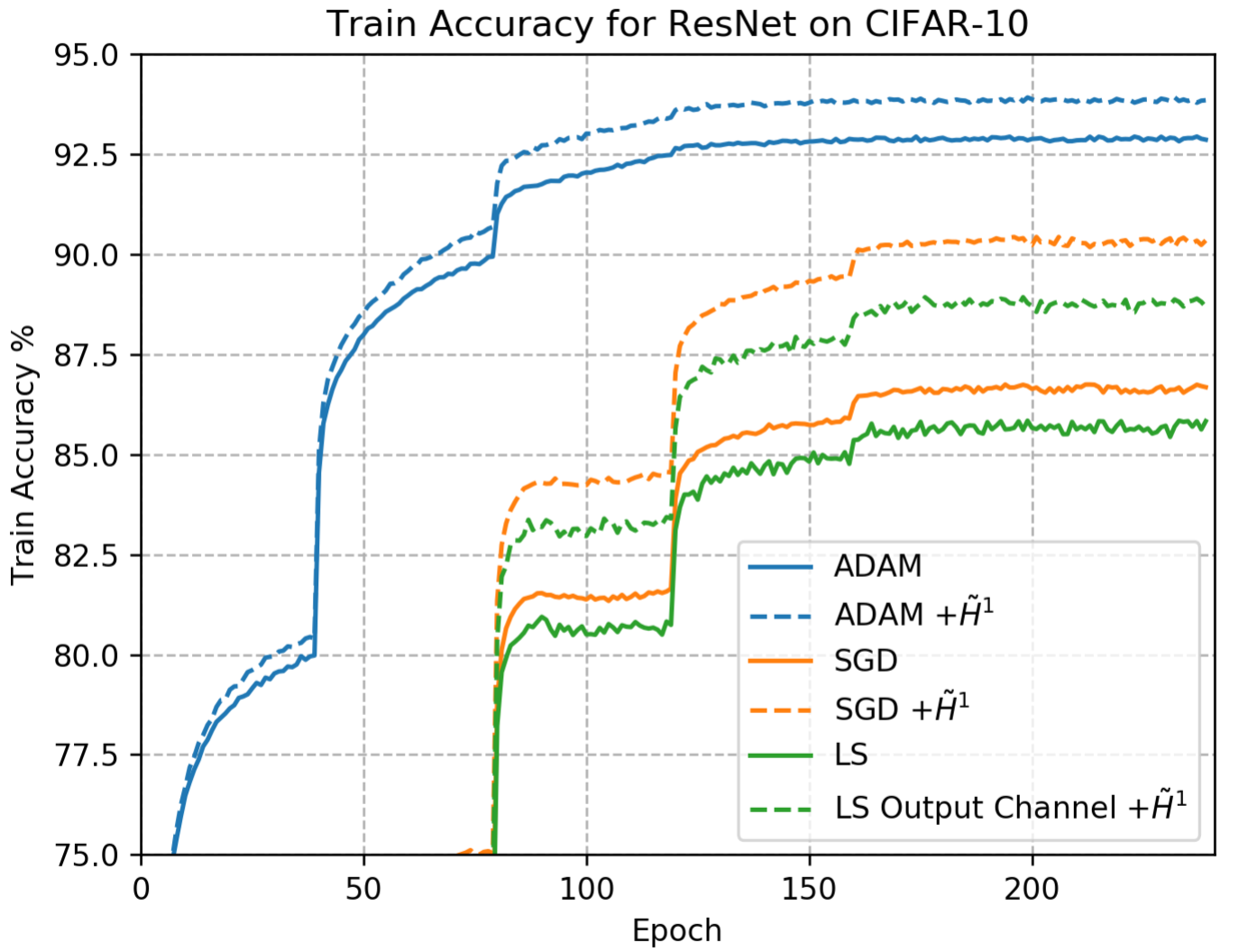}
\endminipage
\minipage{0.45\textwidth}
\includegraphics[height=1.6in]{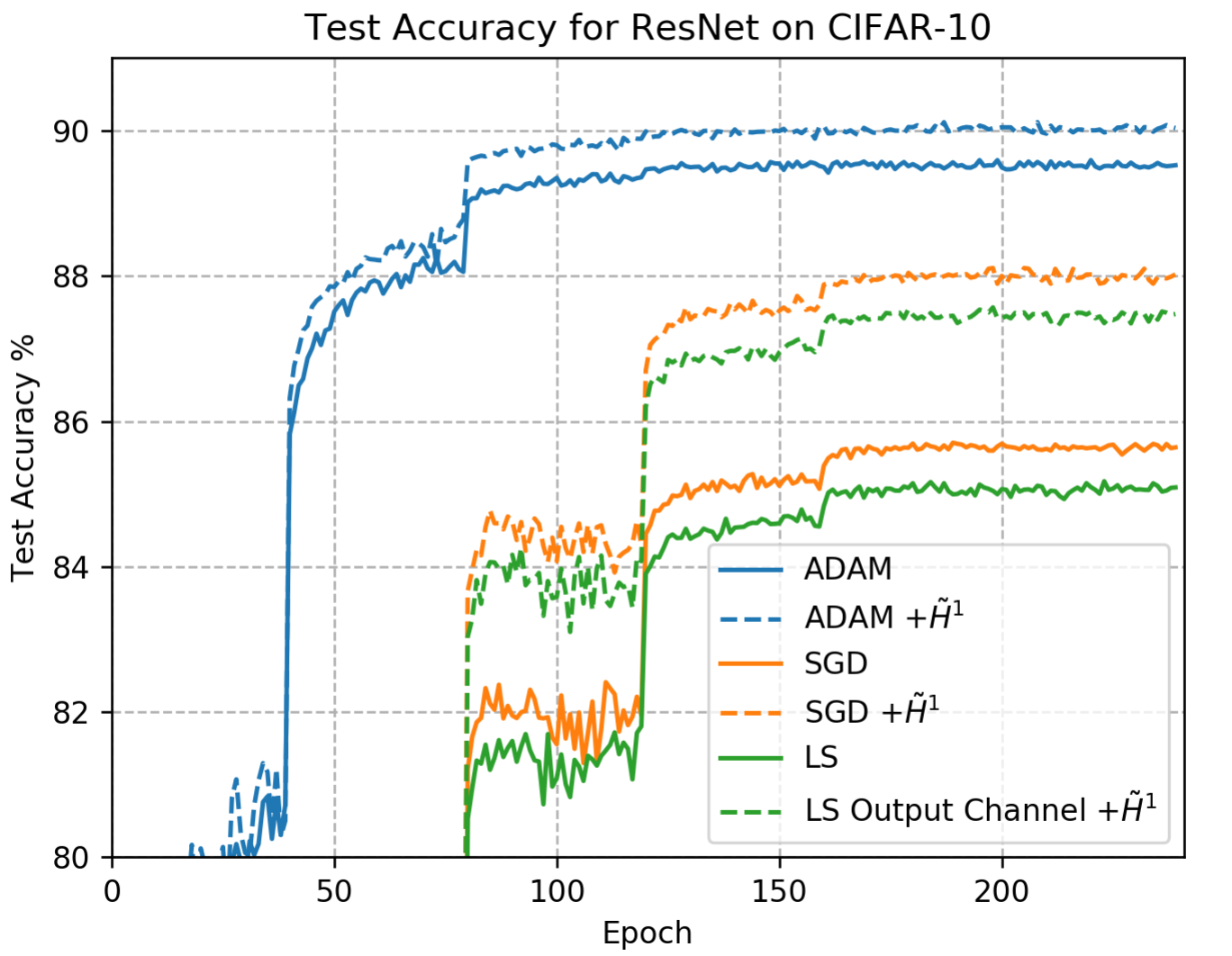}
\endminipage
\vspace{-0.05in}
\caption{{\bf Evolution of training and test accuracy on CIFAR-10: an example with batchsize = 8.} Our metric significantly improves both training and test accuracy. \cut{We also apply smoothing by LS along the same channel direction as ours and achieve improvement over the original LS.}}
  \label{Fig:cifar8}
    \vspace{-0.1in}
\end{figure}

In Figure~\ref{Fig:cifar8}, we show an example of training and test accuracy curves (batch size of 8) for baselines as well as Laplacian Smoothing (LS) \cite{osher2018laplacian}, which rasterizes before smoothing. We out-perform all methods. We also apply LS (without rasterization) to smooth the gradient in our output-channel directed fashion, which improves LS, but we still out-perform it. The original implementation of LS only runs smoothing for the first 40 epochs, then uses SGD (for speed). In our experiments, we apply smoothing all the way to convergence to test the effectiveness for the whole duration. 

\begin{figure}[H]
\centering
\minipage{0.35\textwidth}
\includegraphics[height=1.3in]{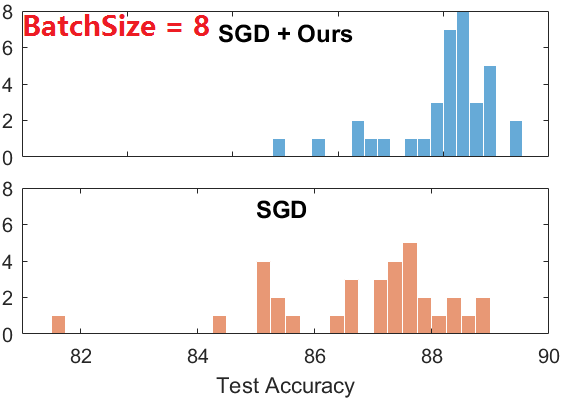}%
\endminipage
\minipage{0.35\textwidth}
\includegraphics[height=1.3in]{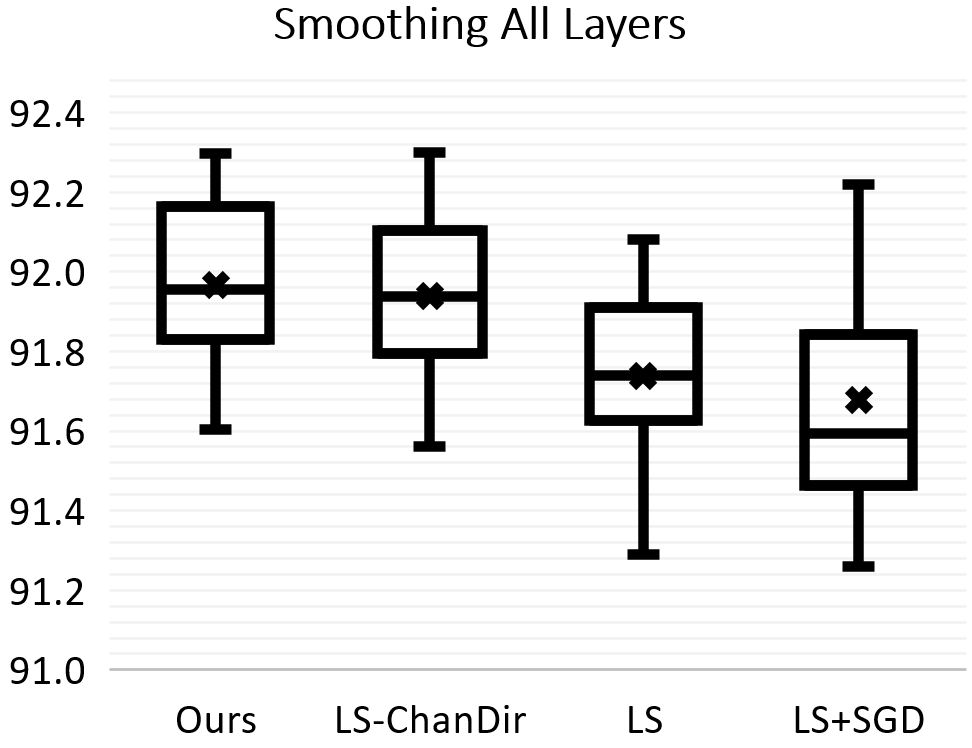}%
\endminipage
\minipage{0.35\textwidth}
\includegraphics[height=1.3in]{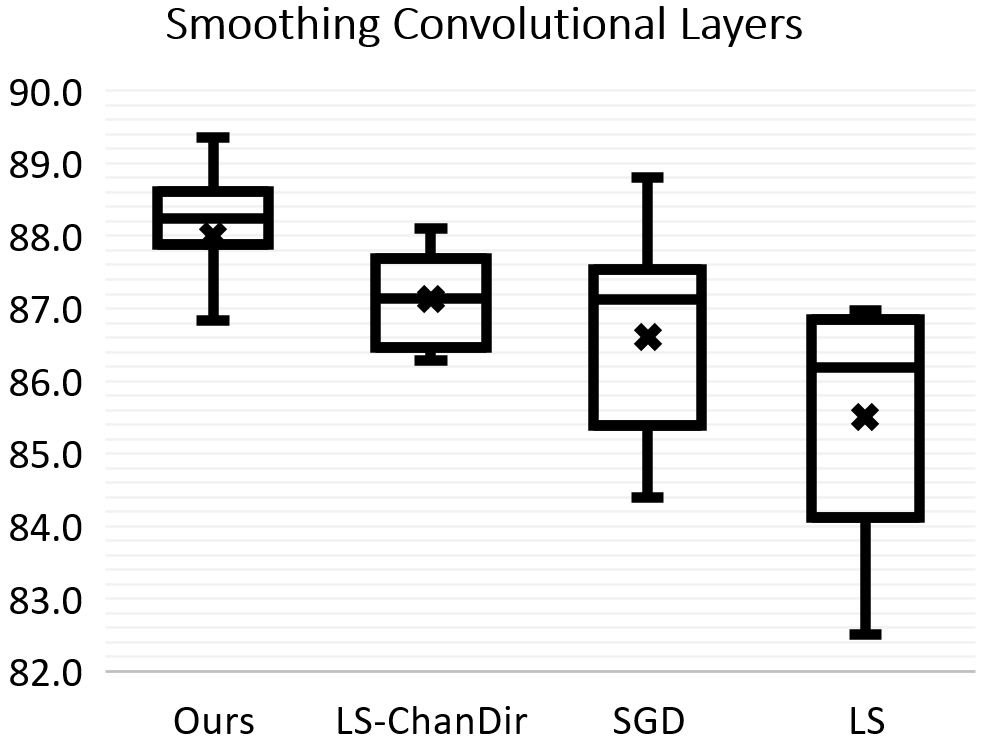}%
\endminipage
\vspace{-0.05in}
\caption{{\bf Distribution of results on CIFAR-10.} {\it Left}: Histogram of test accuracy. Ours achieves higher average with significantly reduced variance. {\it Right}: Results from different smoothing directions. Best accuracy obtained from our proposed direction. {\it A: Output-Channel Directed; B: Input-Channel Directed; All: parameters rasterized into a 1-D vector to perform smoothing; Ours: re-weighted $\mathbb{L}^2$.}}
\label{Fig:cifar_distribution}
  \vspace{-0.1in}
\end{figure}

{\bf Variance reduction}: In Figure~\ref{Fig:cifar_distribution} (left), we compare the histograms of test accuracy over multiple runs of ours and SGD. Our method achieves higher average test accuracy with reduced variance.

{\bf Direction of smoothing}: To investigate the effect of different channel directions of smoothing, we apply our method as well as LS along different channel-directions. We compare approaches under two settings, which are smoothing gradients in all layers and smoothing gradients in only convolutional layers.\footnote{For completeness, we have also tested LS for 40 epochs and then switched to SGD (LS+SGD), as done by \cite{osher2018laplacian} for speed; this results in slightly worse performance than running LS for all epochs. This verifies that running LS for all epochs, as we did in experiments in our main paper gives the best results for LS.} Figure~\ref{Fig:cifar_distribution} (right) shows that our output-channel direction is preferred regardless of smoothing method used. This shows that preferentially treating the output channel to smooth, as in our approach, is essential to performance. Interestingly, smoothing only convolutional layers in a rasterized order (as in LS) performs worse than SGD, but that is made up by smoothing in non-convolutional layers when smoothing in all layers.

\begin{figure}[H]
\centering
\includegraphics[height=1.3in]{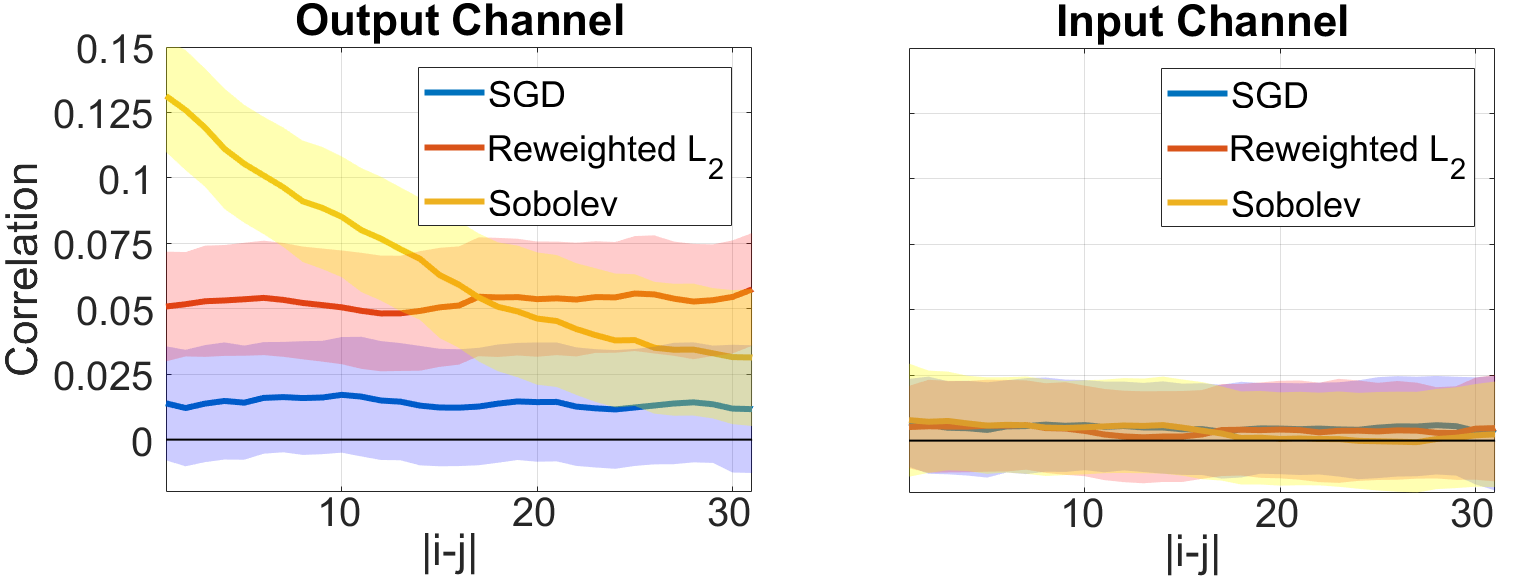}%
\caption{{\bf Regularity of Tensor.} Correlation between weights within different channel directions in CIFAR trained ResNet 56 conv layers (over 10 trials). $|i - j|$ is distance between weight locations in tensor for correlation computation. Sobolev/re-weighted $H^0$ show strong correlation in output direction, but not input. SGD shows correlation in output direction.}
\label{Fig:cifar_regularity}
  \vspace{-0.1in}
\end{figure}

{\bf Regularity of Tensor}: We show that the final weight tensor at convergence in our methods have regularity in the output channel dimension in Figure ~\ref{Fig:cifar_regularity}, as should be the case as the tensor is composed of a component that is smooth. To show this, we plot the correlation between filters in the weight tensors as a function of the distance in the output channel dimension. This is done over multiple tensor layers in ResNet-56 and over multiple trials of optimization on CIFAR-10. We also show the correlation of filters in the input channel direction. As can be seen, all optimization methods produce tensors that exhibit regularity in the output channel direction while no such regularity in the output direction. Notice that our methods increase the amount of regularity compared to SGD as it imposes this in optimization.

\begin{figure}[H]
\vspace{-0.1in}
\centering
\minipage{0.45\textwidth}
\includegraphics[height=1.5in]{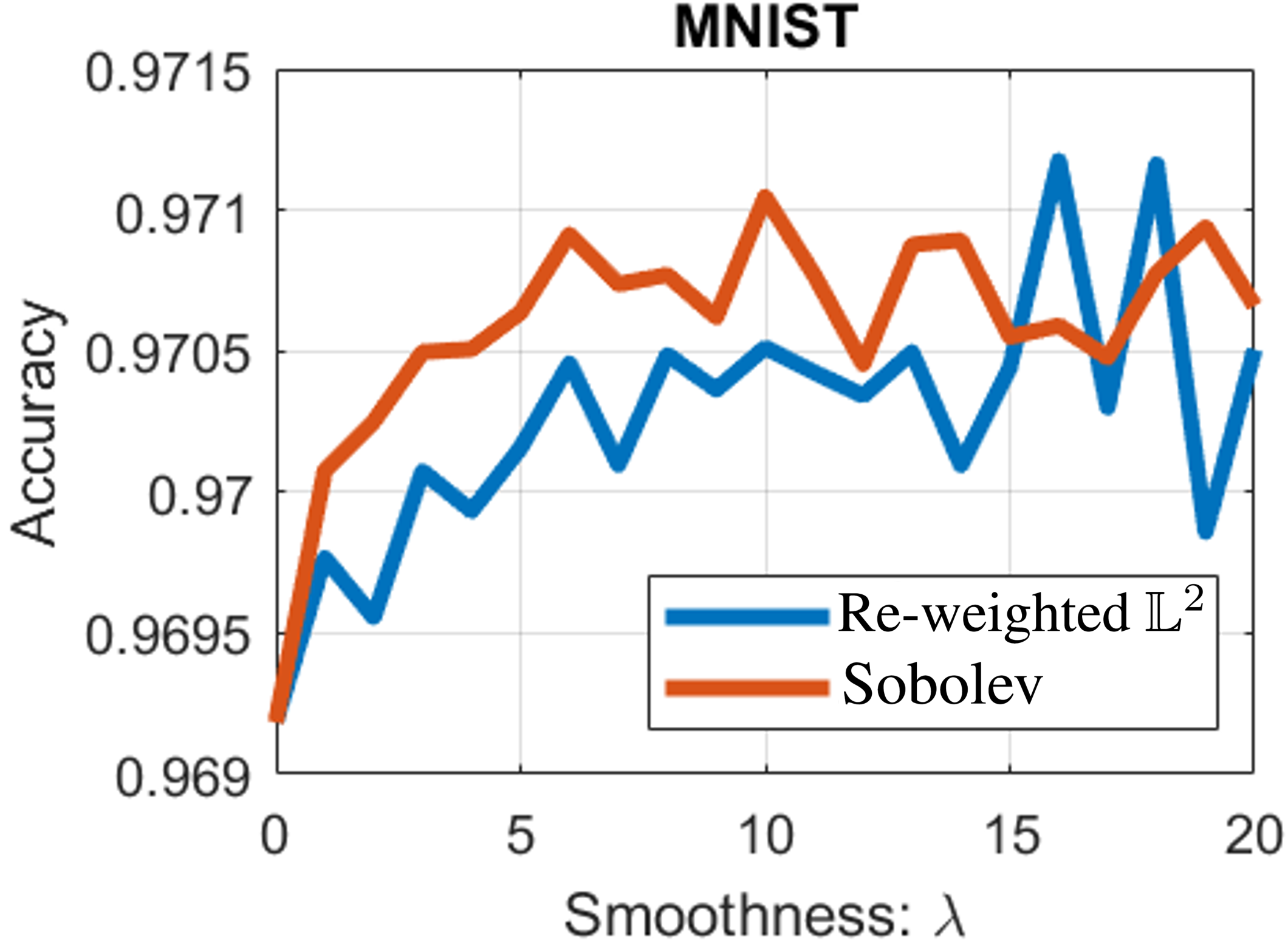}
\endminipage
\minipage{0.4\textwidth}
\includegraphics[height=1.52in]{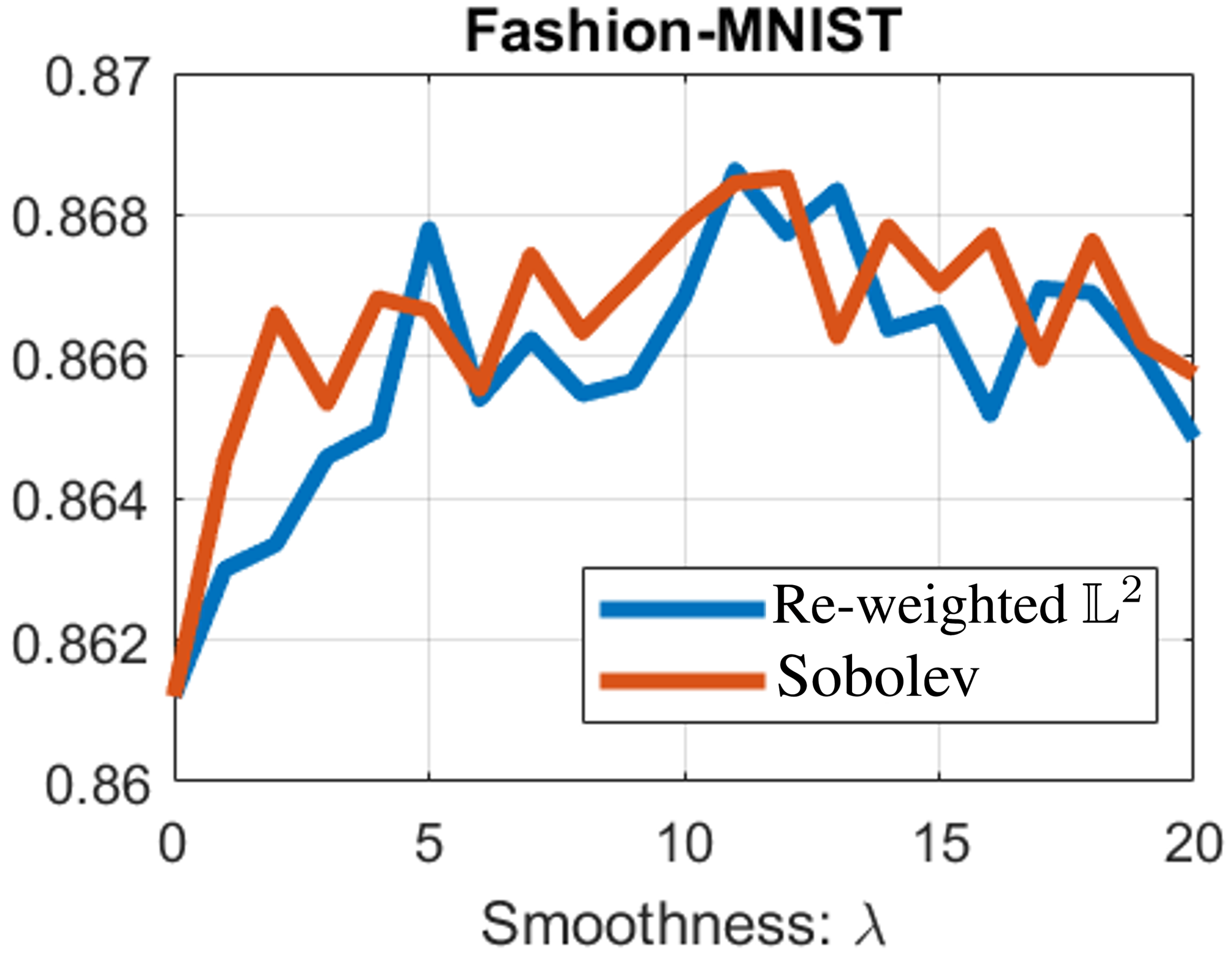}
\endminipage
\vspace{-0.05in}
\caption{{\bf Results on MNIST and Fashion-MNIST with different choice of smoothness.} 0ur methods improve classification accuracy over SGD (when $\lambda=0$) for a wide range of smoothness levels.}
\label{fig:mnist}
  \vspace{-0.1in}
\end{figure}

{\bf Effect of Smoothing Parameter}: We perform controlled experiments on MNIST~\cite{lecun-mnisthandwrittendigit-2010} and Fashion-MNIST~\cite{xiao2017fashionmnist} by varying the smoothness parameter $\lambda$ from 0 to 20. Instead of using the standard partition, we conduct training on the test set (10000 samples) and test on the training set (60000 samples), which makes generalization more challenging. We use a 2-layer CNN with 50 and 100 $5\times5$ filters in each layer, respectively, and train with batch size 100. Figure~\ref{fig:mnist} shows the accuracy at the 100th epoch (average over 5 trials). When $\lambda=0$, the optimizer degenerates to vanilla SGD. Our methods are not sensitive to $\lambda$ and improve over SGD for any $\lambda$.

\begin{figure}[H]
\centering
\minipage{0.45\textwidth}
\includegraphics[height=1.45in]{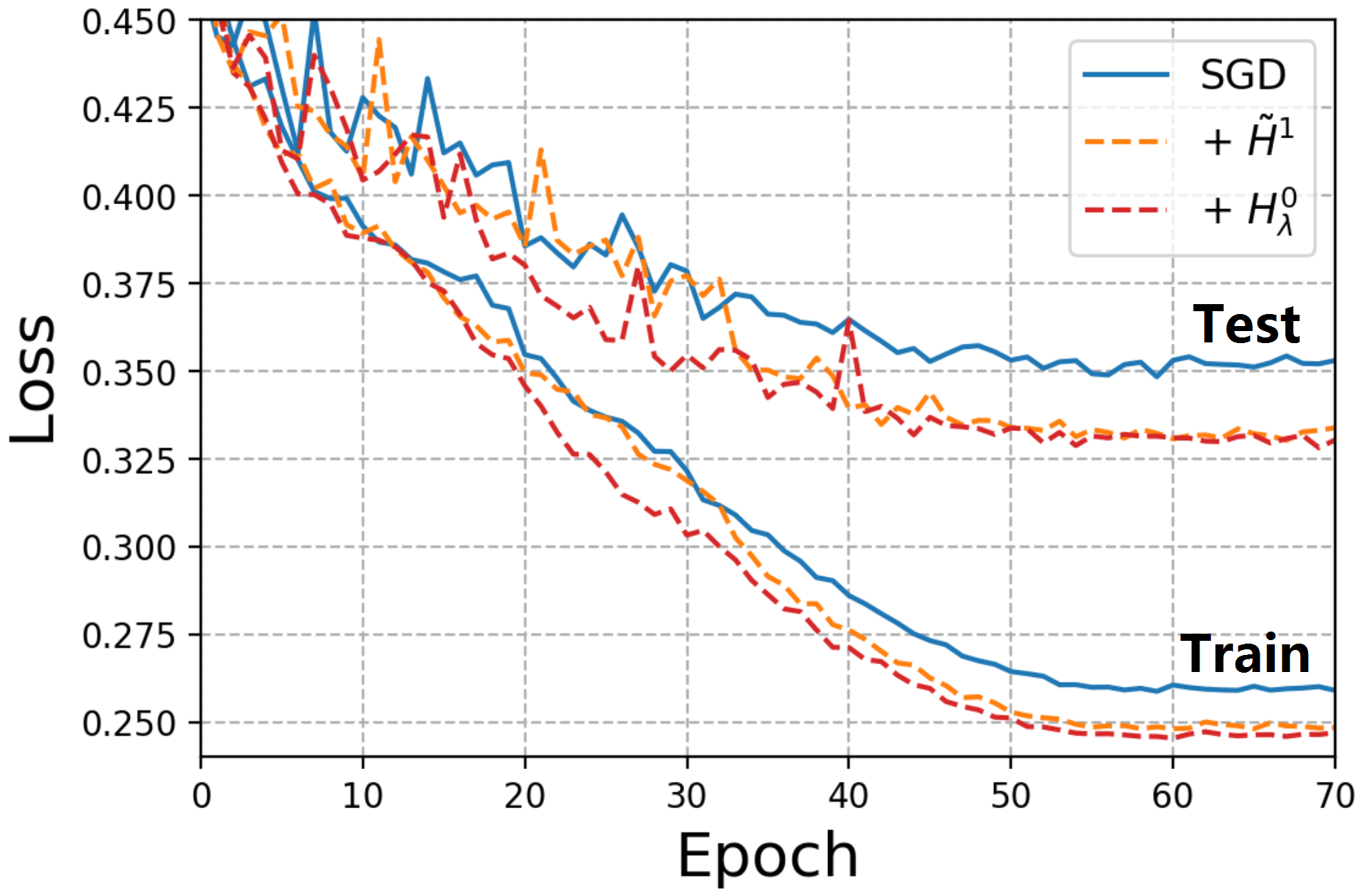}
\endminipage
\minipage{0.4\textwidth}
\includegraphics[height=1.45in]{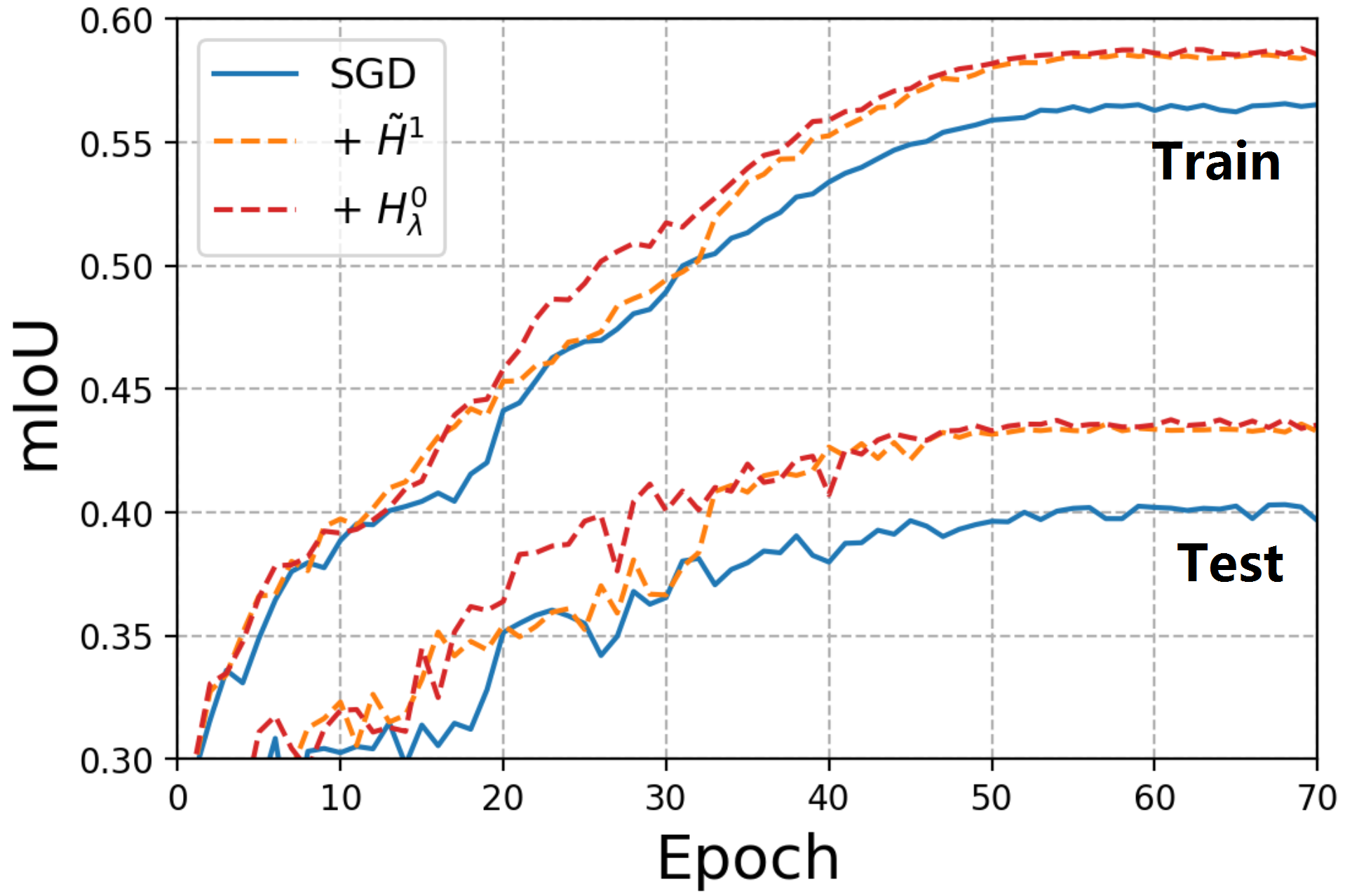}
\endminipage
\vspace{-0.05in}
\caption{{\bf Semantic Segmentation Results on PascalVOC.} Sobolev $\tilde H^1$ and re-weighted $\mathbb{L}^2$ ($H^0_\lambda$) improve segmentation accuracy by 8.5\% and 7.8\% respectively relative to SGD.}
\label{fig:voc}
\vspace{-0.1in}
\end{figure}

{\bf Semantic Segmentation:}
The experiments on semantic segmentation are conducted on the PascalVOC~\cite{Everingham15} dataset using a standard segmentation network \cite{ronneberger2015unet} with ResNet-50 as the encoder (see https://github.com/nyoki-mtl/pytorch-segmentation). We perform training with initial learning rate 7e-3 and batch size 2 (the maximum size to fit on Titan Xp GPU memory), and record the training/testing loss and accuracy for 60 epochs. 3 independent trials are run under each setting. Figure~\ref{fig:voc} shows comparison between ours and SGD. Both channel-directed metrics improve the final segmentation accuracy on the test set by \textasciitilde8\% relatively. Also note that our method reduced the generalization gap from 0.163 to 0.151 (by 7.4\%) and 0.150 (by 8.0\%) for $\tilde H^1$ and $H^0_{\lambda}$, respectively. 

\begin{figure}[H]
\centering
  \begin{minipage}[t]{0.4\textwidth}
    \includegraphics[height=1.6in]{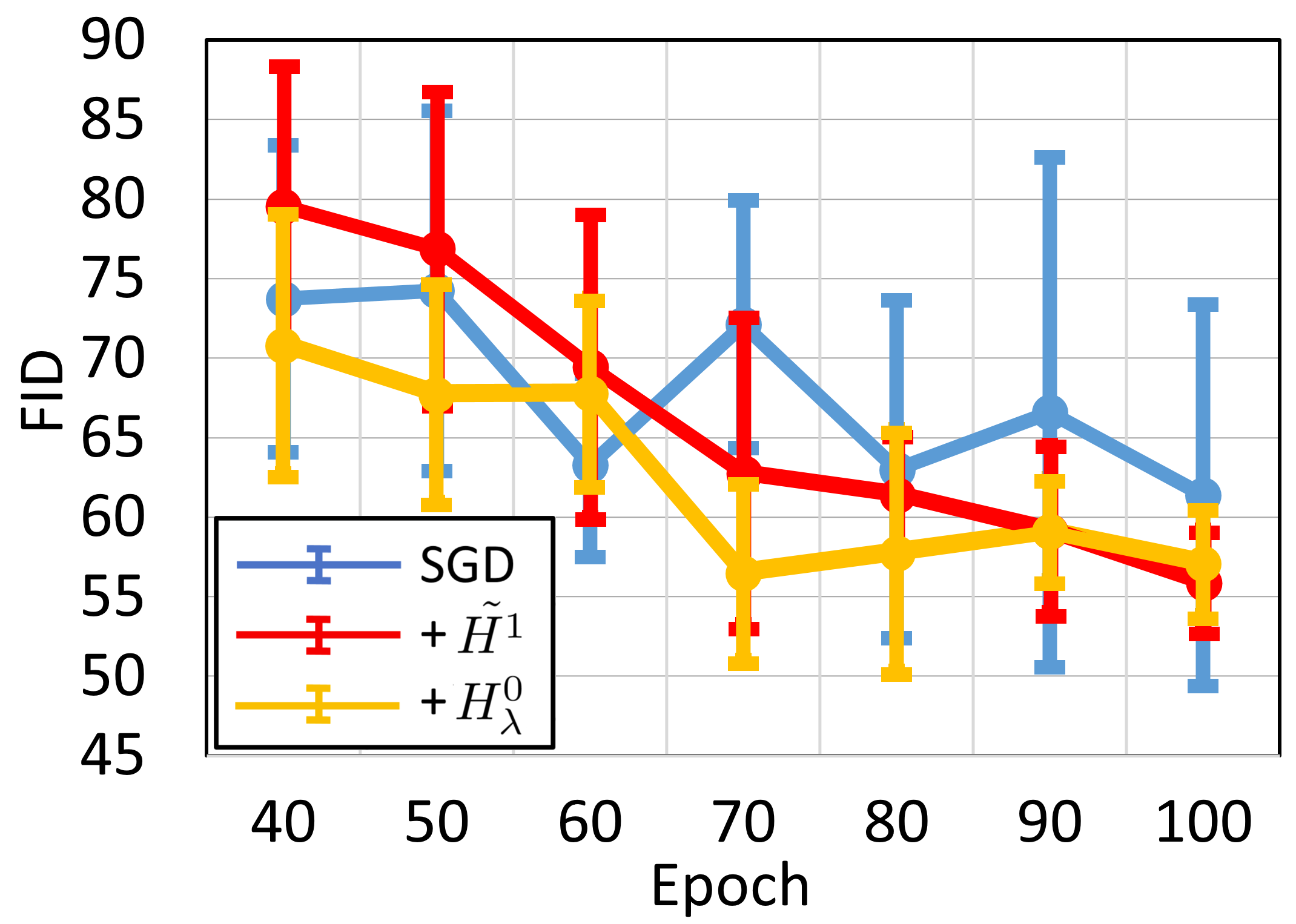}
  \end{minipage}\hfill
  \begin{minipage}[b]{0.55\textwidth}

\caption{{\bf Results on the image generation task.} Our methods achieve better result with significantly reduced variance due to regularity imposed during the training process. {\it Final FID: SGD: $61.37\pm12.00$; Channel-Directed Sobolev ($\tilde{H^1}$): $56.31\pm3.12$; Channel-Directed Re-Weighted $\mathbb{L}^2$ ($H^0_\lambda$): $57.62\pm4.02$. Lower is better.}}\label{fig:gan}
  \end{minipage}

    \vspace{-0.1in}
\end{figure}

{\bf Image Generation:} To test the performance on GANs, we choose the task of semantic labels to image conversion. We perform the experiments on the current state-of-the-art model SPADE~\cite{park2019SPADE} (a.k.a GauGAN), which aims to generate high-quality realistic images from given semantic layouts. Experiments are conducted on CityScapes~\cite{Cordts2016Cityscapes} and the FID~\cite{heusel2017gans} score is used to evaluate the quality (lower is better). Learning rates are $1e-4$ and $4e-4$ for the generator and discriminator, respectively. We compare to SGD with momentum 0.9 and weight decay 5e-4. All models are trained for 100 epochs with batch size 2 (to fit on Tesla v100 memory), and 6 independent trials are run for each optimizer. Figure~\ref{fig:gan} provides FID curves and error bars. Our methods achieve better average FID score with significantly less variance. Note 2 out of 6 models trained by SGD suffered from collapse, which led to high variance, while all twelve trials of our methods achieved good final results.

{\bf Speed:} With PyTorch, re-weighted $\mathbb{L}^2$ ($H^0_\lambda$) adds negligible overhead. In our current Pytorch implementation, $\tilde{H^1}$ adds on average 45 ms overhead to each mini-batch with batch size 128, which increases training time on CIFAR-10 by 50\%. This is because tensor transpose and saving/loading are currently required due to limited library functions, contributing to a large portion of computational overhead. In principle, as computing the $\tilde H^1$ gradient has linear time complexity, if the computation were done, for instance, using C++, it like re-weighted $\mathbb{L}^2$, would add negligible overhead over SGD/Adam.

\section{Conclusion}
Using gradients that are regular in the output-channel dimension of CNN network tensors in SGD is effective in improving generalization accuracy of SGD and its variants. We reformulated the gradient (without changing the loss) by changing the underlying Riemannian geometry on the tensor space using two different metrics. Both the channel-directed re-weighted $H^0$ and $\tilde H^1$ both gave generalization boosts. Regularity in other tensor dimensions was not effective in improving SGD or variants. Both channel-directed gradients have similar computational complexity, and the re-weighted $H^0$ adds negligible training time in its Pytorch implementation, which is one line of PyTorch code.

%\clearpage

\bibliography{main}
\bibliographystyle{iclr2021_conference}

\clearpage
\appendix

\section{Additional Analysis of Evolution of Channel-Directed Optimization}

\begin{figure}[H]
\centering
\minipage{0.45\textwidth}
\includegraphics[height=1.6in]{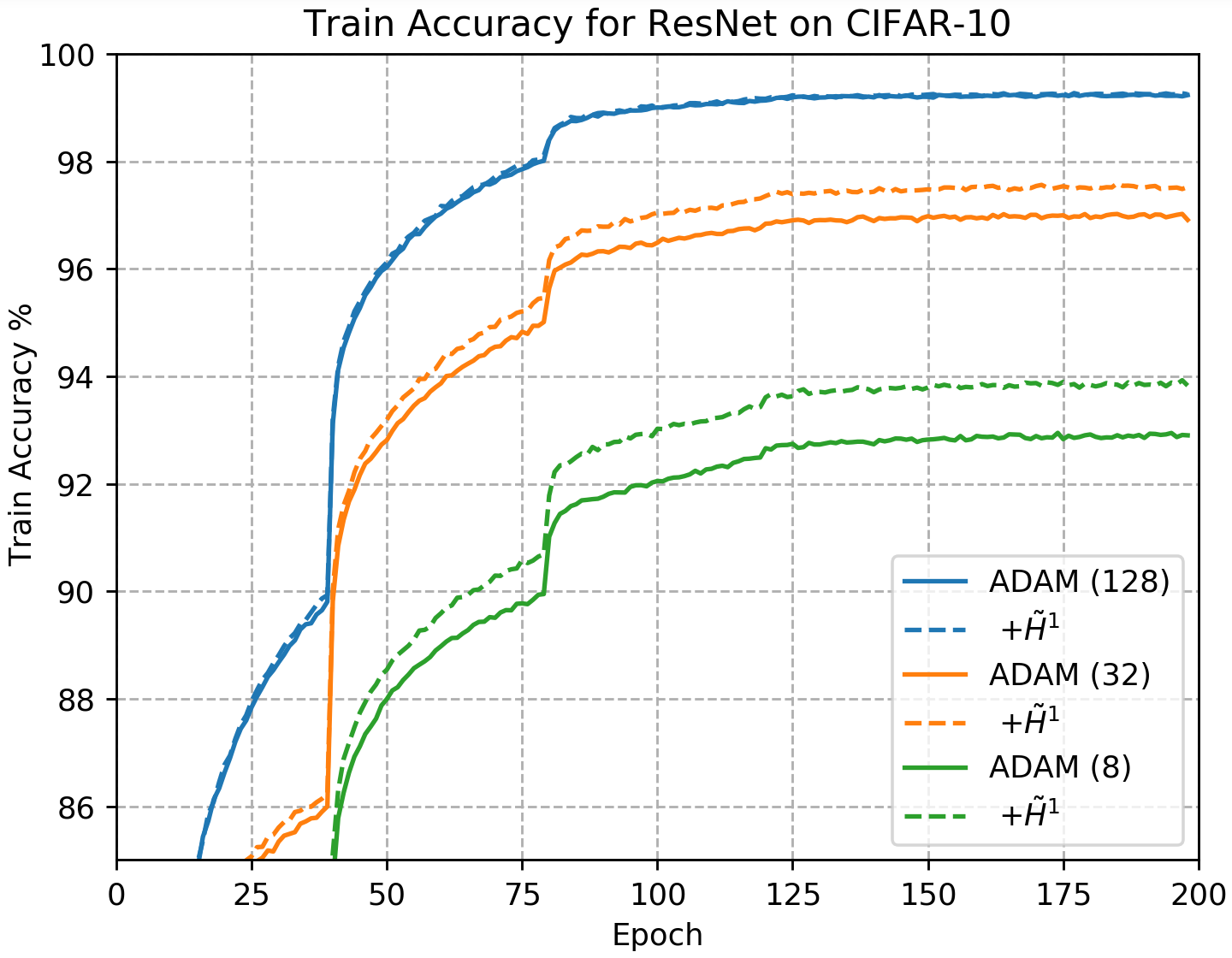}
\endminipage
\minipage{0.45\textwidth}
\includegraphics[height=1.6in]{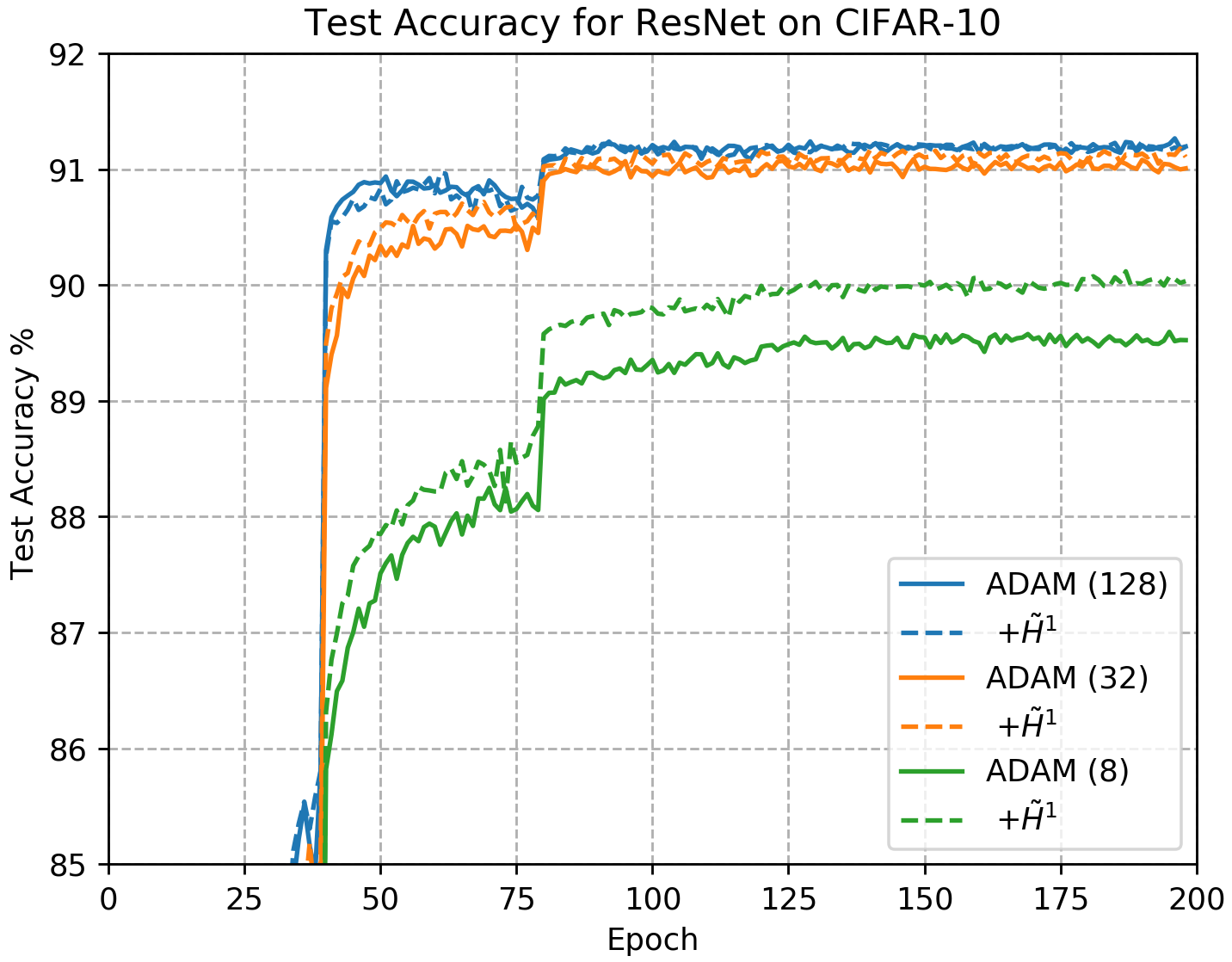}
\endminipage
\caption{{\bf Training and test accuracy on CIFAR-10 with ADAM.}}
  \label{Fig:adam}

\end{figure}

\begin{figure}[H]
\centering
\minipage{0.45\textwidth}
\includegraphics[height=1.6in]{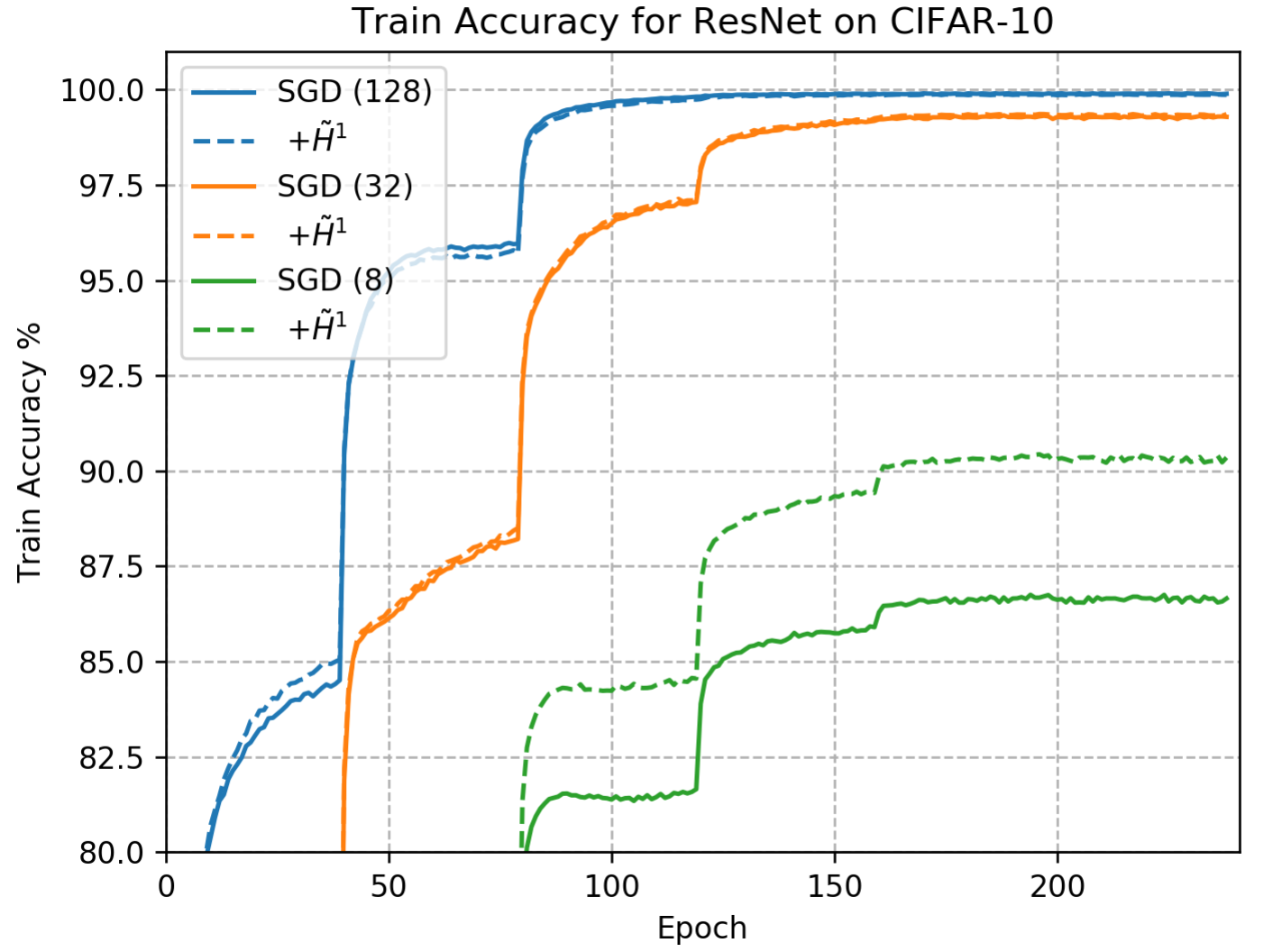}
\endminipage
\minipage{0.45\textwidth}
\includegraphics[height=1.6in]{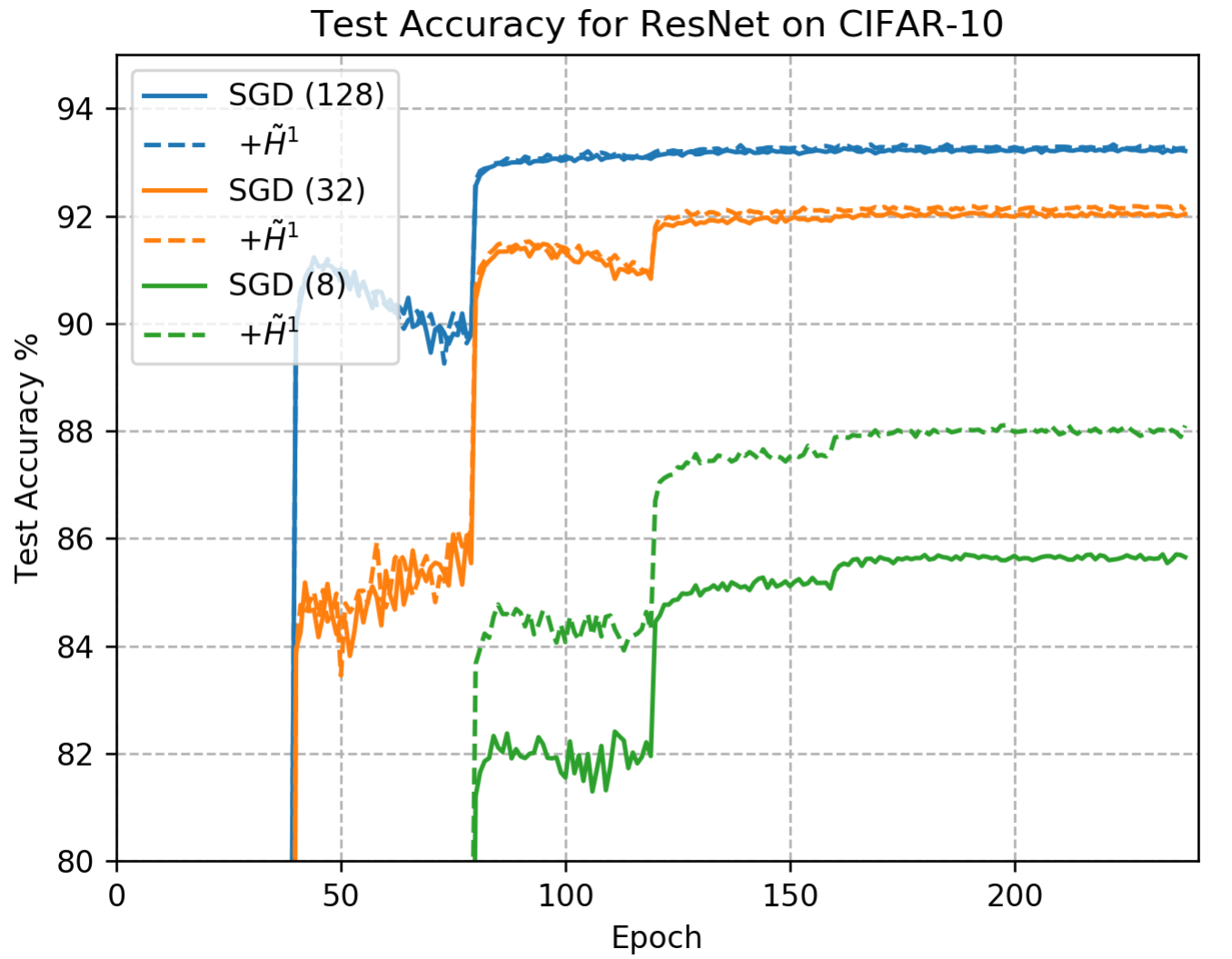}
\endminipage
\caption{{\bf Training and test accuracy on CIFAR-10 with SGD.}}
  \label{Fig:sgd}

\end{figure}

Figure~\ref{Fig:adam} and Figure~\ref{Fig:sgd} present the evolution of training and test accuracy of ADAM and SGD with different batch sizes.  Using channel-directed gradients ($\tilde H^1$ in this experiment) for SGD or ADAM improves test accuracy for any batch size. More prominent performance gains are seen for smaller batch sizes.

\section{Additional Experimental Verification of Output-Channel Direction}
\begin{figure}[H]
\centering
\includegraphics[height=1.6in]{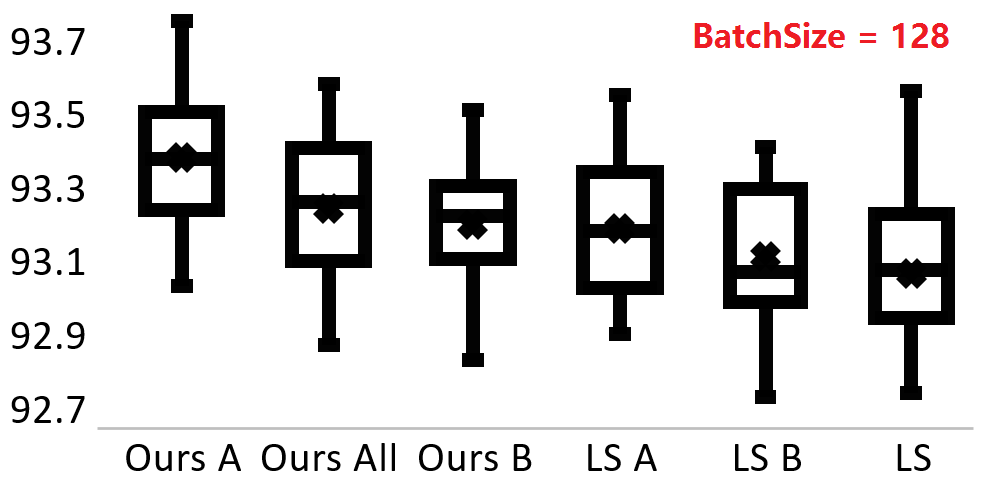}
\caption{{\bf Channel-Directed Smoothing Leads to Better Performance.}  Best accuracy obtained from our proposed direction. {\it A: Output-Channel Directed; B: Input-Channel Directed; All: parameters rasterized into a 1-D vector to perform smoothing; Ours: re-weighted $\mathbb{L}^2$.}}
  \label{Fig:cifar128}
\end{figure}

To investigate the effect of different channel directions of smoothing, we apply our method as well as LS along different channel-directions. Figure~\ref{Fig:cifar128} shows that our output-channel direction is preferred regardless of different smoothing approaches.

\cut{
In this experiment on CIFAR-10 (for batch size 8), we further demonstrate that our output channel-directed gradients are essential in performance by comparing to Laplacian Smoothing (LS) \cite{osher2018laplacian}, which also smooths the gradient, but in a rasterized ordering of components of the gradient, rather than in our channel-directed fashion.

On the left of Figure~\ref{Fig:cifar8}, we only use our channel-directed approach ($\tilde H^1$) on convolutional layers. We compare to applying LS only to convolutional layers and SGD. Our approach performs best, and suprisingly LS performs worse than SGD. We also show that using LS but smoothing in our output-channel directed fashion (LS-ChanOrd), improves LS significantly.  This shows that preferentially treating the output channel to smooth, as in our approach, is essential to performance.

On the right of Figure~\ref{Fig:cifar8}, we compare approaches when smoothing gradients in all layers. Our channel-directed approach performs better than LS, but by smaller margins. If we perform output channel-directed smoothing of convolutional layers in LS (LS-ChanOrd), the results are nearly the same as our approach, again indicating that the channel-directed smoothing of convolutional layers is essential in performance.  
}

\section{Detailed Derivations for Section 2.2}
We first derive the re-weighted  $\mathbb{L}^2$ gradient under $H^0_\lambda$ metric following the same notations from the paper. Consider $f \triangleq \nabla_{H^0} L(X)$ the standard $\mathbb{L}^2$ gradient, and we want to solve for $g \triangleq \nabla_{H^0_\lambda} L(X)$. By \eqref{eq:reweightedH0_metric} and \eqref{eq:gradient} we have 
\begin{align}
    \inner{0}{f}{k} &= \ip{g}{k}{H^0_{\lambda}} \\
&=\inner{0}{\bar{g}}{\bar{k}} + \lambda\inner{0}{g-\bar{g}}{k-\bar{k}}.
\end{align}
After decomposing $f$ and $k$ into 
\begin{equation}
 f = \bar{f}+(f-\bar{f}),\quad k = \bar{k}+(k-\bar{k})   
\end{equation}
and noting the fact that $\inner{0}{\bar{k}}{k-\bar{k}}=0$ holds for all $k$, we have
\begin{align}
\bar{f} = \bar{g}, \quad f-\bar{f}=\lambda(g-\bar{g}), 
\end{align}
which leads to the result of \eqref{eq:reweightedH0_grad}.

We then derive the Sobolev gradient under $H^1$ metric, following similar computations in \cite{sundaramoorthi2007sobolev}. Consider $\nabla_{H^1} L(X)$ the Sobolev gradient under $H^1$ metric. By \eqref{eq:metric_H1} and \eqref{eq:gradient} we have 
\begin{align}
    \inner{0}{\nabla_{H^0} L(X)}{k} &= \inner{1}{\nabla_{H^1} L(X)}{k} \\
&=\frac{1}{O}\inner{0}{k}{\nabla_{H^1} L(X)} + \lambda O \inner{0}{\pder{k}{o}}{\pder{\nabla_{H^1} L(X)}{o}}.
\end{align}
Integrating by parts and considering the periodic boundary conditions, we have
\begin{equation}
\inner{0}{\nabla_{H^0} L(X)}{k} = \inner{0}{\nabla_{H^1} L(X) - \lambda O^2 \pder{^2}{o^2} \nabla_{H^1} L(X)}{k}.
\end{equation}
Since $k$ can be any perturbation, by uniqueness, we have 
\begin{equation}
\nabla_{H^0}L(X) = \nabla_{H^1} L(X) - \lambda O^2 \pder{^2}{o^2} \nabla_{H^1} L(X)
\end{equation}
which is \eqref{eq:H1_grad}. Similarly, for $\tilde{H^1}$ metric, we have 
\begin{equation}
\label{eq:H}
\nabla_{H^0}L(X) =\avg{ \nabla_{\tilde H^1} L(X)} - \lambda O^2 \pder{^2}{o^2} \nabla_{\tilde H^1} L(X).
\end{equation}

First observe that by computed the output-channel directed average of the both sides of the above equation, we see that $\overline{\nabla_{\tilde H^1} L(X)} = \overline{\nabla_{H^0} L(X)}$, i.e., the average values are same. One may integrate \eqref{eq:H} twice to solve for the $\tilde{H^1}$ gradient. For simplicity, let $f$ be the $\mathbb{L}^2$ gradient and $g$ be the $\tilde{H^1}$ gradient. Integrating twice yields 
\begin{align}
    g(o,i,h,w) &= g(0,i,h,w) + \int_0^{o} \pder{g}{o}(0,i,h,w) \ud \tilde o - \frac{1}{\lambda} \int_0^{o}\int_0^{\hat{o}}(f(\tilde oO,i,h,w)-\bar{f}(i,h,w)) \ud \tilde o \ud \hat{o}\\
    &= g(0,i,h,w) + \int_0^{o} \pder{g}{o}(0,i,h,w) \ud \tilde o - \frac{1}{\lambda} \int_0^{o}\int_{\tilde o}^{o}(f(\tilde oO,i,h,w)-\bar{f}(i,h,w)) \ud \hat{o} \ud \tilde o\\
    &= g(0,i,h,w) + o \pder{g}{o}(0,i,h,w) - \frac{1}{\lambda} \int_0^{o}(o-\tilde o)(f(\tilde oO,i,h,w)-\bar{f}(i,h,w)) \ud \tilde o. \label{eq:th1_sln}
\end{align}
Note that here we perform normalization by scaling to the channel direction by letting $o\in [0,1]$. With boundary conditions $g(0) = g(1)$, $\pder{g}{o}(0) = \pder{g}{o}(1)$ and $\bar{f} = \bar{g}$, we have
\begin{equation}
    \pder{g}{o}(0,i,h,w) = -\frac{1}{\lambda} \int_0^1 o(f(oO,i,h,w) - \bar{f}(i,h,w) ) \ud o. 
\end{equation}
For simplicity, we eliminate $i,h,w$ and $O$ in the following derivations. We have
\begin{align}
    \label{eq:integral}
    g(0) &= g(o) - o\pder{g}{o}(0) + \frac{1}{\lambda}\int_0^{o}(o-\tilde o)(f(\tilde o)-\bar{f}) \ud \tilde o\\
    &= g(o) + o\frac{1}{\lambda} \int_0^1 o(f(o) - \bar{f}) \ud o + \frac{1}{\lambda}\int_0^{o}(o-\tilde o)(f(\tilde o)-\bar{f}) \ud \tilde o .
\end{align}
Noting $\int_0^1 g(0) \ud o = g(0)$ and $\bar{f} = \int_0^1 f(o) \ud o$, we integrate both sides over the entire interval $[0,1]$. 
\begin{align}
    g(0) &= \bar{g} + \frac{1}{\lambda}\int_0^1 o \ud o \cdot \int_0^1 o(f(o) - \bar{f}) \ud o +\frac{1}{\lambda}\int_0^1\int_0^{o}(o-\tilde o)(f(\tilde o)-\bar{f}) \ud \tilde o \ud o\\
    &= \bar{f} + \frac{1}{2\lambda} \int_0^1 of(o) \ud o - \frac{1}{4\lambda}\bar{f} +\frac{1}{\lambda}(\int_0^1\int_0^{o}(o -\tilde o )f(\tilde o)\ud \tilde o \ud o + \bar{f} \int_0^1\int_0^{o}(o -\tilde o)\ud \tilde o \ud o \\
    &=(1-\frac{1}{4\lambda}-\frac{1}{6\lambda})\bar{f}+\frac{1}{2\lambda} \int_0^1 of(o) \ud o + \frac{1}{\lambda}\int_0^1\int_{\tilde o}^1(o -\tilde o )f(\tilde o)\ud o \ud \tilde o \\
    &=(1-\frac{5}{12\lambda})\int_0^1 f(o) \ud o+\frac{1}{2\lambda} \int_0^1 of(o) \ud o + \frac{1}{\lambda}\int_0^1(\frac{1}{2} + \frac{{\tilde o}^2}{2} - \tilde o)f(\tilde o)\ud \tilde o\\
    &=\int_0^1 (1 + \frac{o^2-o+1/6}{2\lambda})f(o) \ud o.
\end{align}
This gives \eqref{eq:tildeH1_linear_soln_3} in the main paper.

\end{document}